\documentclass[10pt,twocolumn,letterpaper]{article}

\usepackage{cvpr}              %

\usepackage[dvipsnames]{xcolor}
\usepackage[utf8]{inputenc}
\usepackage{times}
\usepackage{epsfig}
\usepackage{graphicx}
\usepackage{amsmath,mathtools}
\usepackage{amssymb}
\usepackage{amsthm}
\usepackage{nicefrac}       %
\usepackage{microtype}      %
\usepackage{amsmath} 
\usepackage{float}
\usepackage{amsfonts}
\usepackage{bm}
\usepackage{enumitem}
\usepackage{mathtools}
\usepackage{multirow}
\usepackage{siunitx}
\usepackage[T1]{fontenc}

\usepackage{fontawesome}
\usepackage{pifont}
\usepackage{color}
\usepackage[algo2e,inoutnumbered,linesnumbered,algoruled,vlined]{algorithm2e}
\usepackage{algpseudocode}
\usepackage{booktabs}
\usepackage{tabularx}
\usepackage{bm}
\usepackage{url}
\usepackage{bbm}
\usepackage{threeparttable}
\usepackage{tikz}
\usepackage{wrapfig}
\usetikzlibrary{backgrounds}
\usetikzlibrary{decorations.pathreplacing}
\usepackage[outline]{contour}
\usepackage{appendix}

\definecolor{cvprblue}{rgb}{0.21,0.49,0.74}
\usepackage[pagebackref,breaklinks,colorlinks,citecolor=cvprblue]{hyperref}

\usepackage[accsupp]{axessibility}  %

\newcommand{\LongName}{$\mathcal{X}^3$\xspace}
\newcommand{\ShortName}{XCube\xspace}

\newcommand{\parahead}[1]{\vspace{1mm}\noindent\textbf{#1.}\ }

\theoremstyle{definition}

\newcommand{\RR}{\mathbb{R}}

\newcommand{\Gauss}{{\mathcal{N}}}
\newcommand{\Zero}{\mathbf{0}}
\newcommand{\Identity}{{\mathbf{I}}}
\newcommand{\loss}{\mathcal{L}}

\DeclarePairedDelimiterX{\infdivx}[2]{(}{)}{%
  #1\;\delimsize\|\;#2%
}
\newcommand{\infdiv}{\mathbb{KL}\infdivx}

\definecolor{darkblue}{RGB}{49,130,189}
\definecolor{stanfordgrey}{RGB}{46,45,41}
\definecolor{cardinalred}{RGB}{253,141,60}

\newcommand{\GridSet}{{\mathcal{G}}}
\newcommand{\Grid}{{\mathbf{G}}}
\newcommand{\AttrSet}{{\mathcal{A}}}
\newcommand{\Attr}{\mathbf{A}}
\newcommand{\LatentSet}{{\mathcal{X}}}
\newcommand{\Latent}{{\mathbf{X}}}
\newcommand{\Condition}{{\mathbf{C}}}
\newcommand{\encw}{{\bm{\phi}}}     %
\newcommand{\decw}{{\bm{\psi}}}     %
\newcommand{\dmw}{{\bm{\theta}}}      %
\renewcommand{\dots}{...}
\newcommand{\expec}{\mathbb{E}}
\newcommand{\netmu}{{\bm{\mu}}}
\newcommand{\netv}{{\bm{v}}}

\newcommand{\change}[1]{{#1}}

\usepackage[capitalize]{cleveref}

\crefname{section}{\S}{\S\S}
\crefname{subsection}{\S}{\S\S}
\crefname{conj}{Conj.}{Conj.}

\Crefname{assumption}{\textbf{H}\hspace{-3pt}}{\textbf{H}\hspace{-3pt}}
\crefname{assumption}{\textbf{H}}{\textbf{H}}

\crefname{algorithm}{\text{Alg.}}{\text{Alg.}}
\crefname{assumption}{\textbf{H}}{\textbf{H}}
\crefname{equation}{\text{Eq}}{\text{Eq}}
\crefname{definition}{\text{Dfn.}}{\text{Dfn.}}
\crefname{lemma}{\text{Lemma}}{\text{Lemma}}
\crefname{dfn}{\text{Dfn.}}{\text{Dfn.}}
\crefname{thm}{\text{Thm.}}{\text{Thm.}}
\crefname{tab}{\text{Tab.}}{\text{Tab.}}
\crefname{fig}{\text{Fig.}}{\text{Fig.}}
\crefname{table}{\text{Tab.}}{\text{Tab.}}
\crefname{figure}{\text{Fig.}}{\text{Fig.}}

\definecolor{mygreen}{RGB}{159, 200, 59}
\definecolor{myred}{RGB}{223, 135, 102}

\def\paperID{613} %
\def\confName{CVPR}
\def\confYear{2024}

\title{\change{\ShortName}: Large-Scale 3D Generative Modeling using Sparse Voxel Hierarchies}
\author{Xuanchi Ren$^{1,2,3}$ \;\; Jiahui Huang$^1$ \;\; Xiaohui Zeng$^{1,2,3}$ \;\; Ken Museth$^1$ \\ Sanja Fidler$^{1,2,3}$ \;\; Francis Williams$^1$\\
\small $^1$NVIDIA \quad $^2$University of Toronto \quad $^3$Vector Institute
}

\begin{document}
\twocolumn[{%
\renewcommand\twocolumn[1][]{#1}%
\maketitle
\begin{center}
\vspace{-2.05 em}
\renewcommand\arraystretch{0.5} 
\centering

\includegraphics[width=\linewidth]{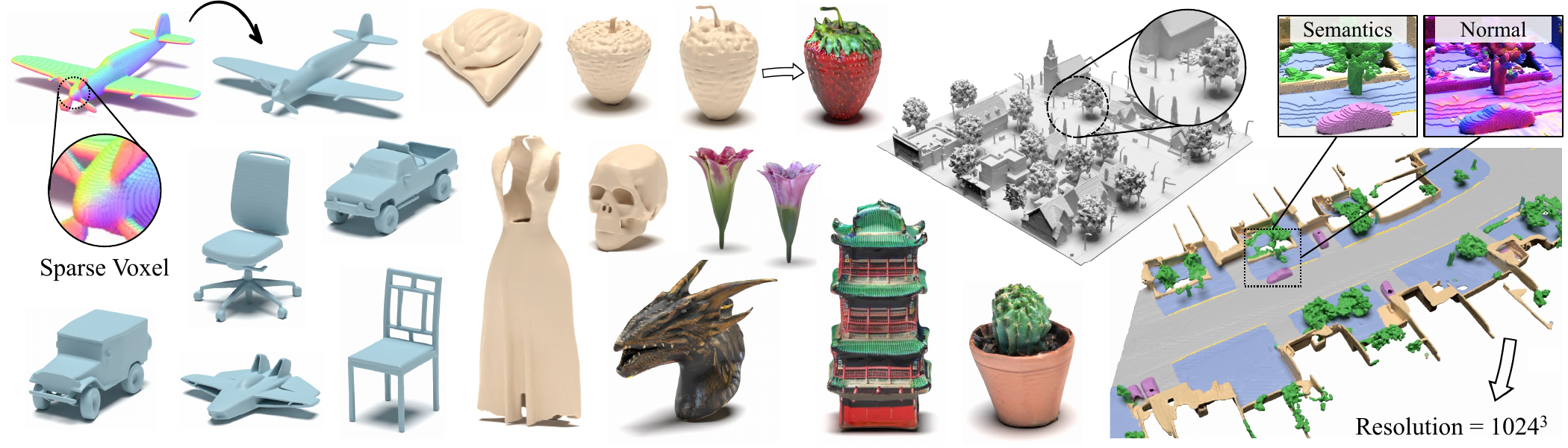}
\vspace{-6mm}
\captionof{figure}{\textbf{\ShortName (\LongName).} Our model generates high-resolution (up to $1024^3$) sparse 3D voxel hierarchies of \emph{objects} and \emph{driving scenes} in under 30 seconds. The voxels are enriched with arbitrary attributes such as semantics, normals, and TSDF from which mesh could be readily extracted. Here we show randomly sampled geometries using our model trained on ShapeNet, Objaverse, Karton City, and Waymo.}
\label{fig:teaser}
\end{center}
}]

\begin{abstract}
We present \change{\ShortName (abbreviated as \LongName)}, a novel generative model for high-resolution sparse 3D voxel grids with arbitrary attributes. Our model can generate millions of voxels with a finest effective resolution of up to $1024^3$ in a feed-forward fashion without time-consuming test-time optimization. 
To achieve this, we employ a hierarchical voxel latent diffusion model which generates progressively higher resolution grids in a coarse-to-fine manner using a custom framework built on the highly efficient VDB data structure. 
Apart from generating high-resolution objects, we demonstrate the effectiveness of \ShortName on large outdoor scenes at scales of \SI{100}{\meter}$\times$\SI{100}{\meter}
 with a voxel size as small as \SI{10}{\centi\meter}.
 We observe clear qualitative and quantitative improvements over past approaches. In addition to unconditional generation, we show that our model can be used to solve a variety of tasks such as user-guided editing, scene completion from a single scan, and text-to-3D. 
 The source code and more results can be found \href{https://research.nvidia.com/labs/toronto-ai/xcube/}{on our project webpage}.
\end{abstract}

\vspace{-4mm}
\section{Introduction}
\label{sec:introduction}

Equipping machines with the ability to create and understand three-dimensional scenes and objects has long been a tantalizing pursuit, promising a bridge between the digital and physical worlds. The problem of 3D content generation lies at the heart of this endeavor. By modeling the distribution of objects and scenes, generative models can propose plausible shapes and their attributes from scratch, from user input, or from partial observations. 

There has been a surge of new and exciting works on generative models for 3D shapes in recent years. Initial work in this area, train on datasets of 3D shapes and leverage 3D priors to perform generation~\cite{NWD, gupta20233dgen}. While these works produce impressive results, their diversity and shape quality is fundamentally bounded by the size of 3D datasets (\eg \cite{Shapenet,Objaverse}), as well as their underlying 3D representation. 
To address the diversity problem, one line of work \cite{poole2022dreamfusion,wang2023prolificdreamer} proposed an elegant solution that leverages powerful 2D generative models to produce 3D structures using inverse rendering and a diffusion score distillation formulation. 
While they benefit from the massive corpora of 2D image data and can generate highly diverse and high-quality shapes, the generated shapes usually suffer from the Janus face problem, and the generation process requires test-time optimization that is lengthy and computationally expensive.
More recent works such as \cite{liu2023zero1to3,liu2023one2345} achieve state-of-the-art 3D generation by smartly combining a mix of 2D and 3D priors. They gain diversity from 2D data and spatial consistency from 3D data and speed up the generation process by operating directly in 3D. 
These works motivate a deeper investigation into the fundamentals of \emph{3D priors} for generation. 
In particular, \change{current 3D priors are limited to low resolutions~\cite{lee2023diffusion}}, and do not scale well to large outdoor scenes such as those in autonomous driving and robotics applications. 
These datasets of large-scale scenes are abundant and contain more data than those consisting solely of objects \cite{waymo}. 
Thus, developing a scalable 3D generative prior has the potential to unlock new sources of training data and further push the boundary of what is possible in 3D generation. 
In this work, we aim to explore the limits of \emph{purely 3D} generative priors, scaling them to high resolutions and large-scale scenes. Our model is capable of scaling to high-resolution outputs (e.g. $1024^3$) by leveraging a novel sparse formulation and can produce outputs with high geometric complexity, by focusing dense geometry near the surface of a shape.

Our method, \LongName, is a novel hierarchical voxel latent diffusion model for generating high-resolution 3D objects and scenes with arbitrary attributes such as signed distances, normals, and semantics. Our model generates a latent \emph{Sparse Voxel Hierarchy} --- a hierarchy of 3D sparse voxel grids with latent features at each voxel --- in a coarse-to-fine manner. In particular, we model each level of the hierarchy as a latent diffusion model conditioned on the coarser level. The latent space at each level is encoded using a highly efficient --- both in terms of compute and memory --- sparse structure Variational Autoencoder (VAE).
Our generated representation enjoys several key benefits: (1) it is fully 3D, enabling it to model intricate details at multiple resolutions, (2) it can output very high-resolution shapes (up to $1024^3$ resolution) by leveraging sparsity, (3) the distribution at each level is easy to model since the coarse level need only model a rough shape, and finer levels are concerned with local details, (4) our generated shapes support multi-scale user-guided editing by modifying coarser levels and regenerating finer levels, (5) since our model leverages a latent diffusion model over a hierarchy of features, we are able to decode \emph{arbitrary} multi-scale attributes (\eg semantics, TSDF) from those features.

We demonstrate the effectiveness of our hierarchical voxel latent diffusion model on standard object datasets such as Objaverse \cite{Objaverse} and ShapeNet \cite{Shapenet} achieving state-of-the-art results on unconditional and conditional generation from texts and category labels. 
We further demonstrate the scalability of our method by demonstrating high-quality unconditional and conditional (from a single lidar scan) generation on large outdoor scenes from the Waymo Open Dataset \cite{waymo} and Karton City \cite{kartoncity}. 
Finally, by leveraging a custom sparse 3D deep learning framework built on VDB~\cite{VDB}, our model is capable of generating complex shapes at $1024^3$ resolution containing \emph{millions of voxels} in under 30 seconds.

\vspace{-1mm}
\section{Related Work}
\label{sec:related}
\vspace{-1mm}

\parahead{Generative Probabilistic Models}%
Common generative models include variational autoencoders (VAE)~\cite{kingma2013auto}, generative adversarial networks (GAN)~\cite{goodfellow2014generative}, normalizing flows~\cite{rezende2015variational}, autoregressive models (AR)~\cite{van2016conditional}, and more recently diffusion models (DM) \cite{sohldickstein2015deep,ho2020denoising}. 
A popular method for generative modeling is \emph{latent diffusion} that has been found useful in, \eg, images~\cite{vahdat2021score,rombach2022highresolution,podell2023sdxl} and videos~\cite{blattmann2023align}, where the diffusion process happens over the latent space of a simpler generative model (\eg a VAE). 
Latent diffusion models allow for easy decoding of multiple attributes through different decoders.
In our work, we employ a latent diffusion model over a hierarchy of sparse voxels. 

\parahead{3D Generative Models}%
The recent surge in the 3D generative modeling literature mostly focuses on \emph{object-level} shape synthesis. 
One line of work opts for distilling 2D image priors into 3D via inverse rendering~\cite{lin2023magic3d,wang2023prolificdreamer,liu2023zero1to3}, while others~\cite{nam20223dldm,shape-e,gao2022get3d,zeng2022lion,ntavelis2023autodecoding,point-e} focus on directly learning 3D priors from large-scale 3D datasets~\cite{Objaverse,deitke2023objaverse}.
Recently, hybrid 2D-3D approaches~\cite{liu2023one,szymanowicz2023viewset} which better leverage both image  priors and large 3D datasets have started to emerge. 
Fundamental to these works are good 3D priors that can instill multiview consistency without the need for expensive test-time optimization. %
This is the interest of our work. 

Works that tackle large-scale \emph{scene} generation either choose a procedural approach that decouples the generation of different scene components (\eg roads, buildings, etc.)~\cite{xie2023citydreamer,infinigen2023infinite,sun20233d}, or a more generic approach that generates the entire scene at once~\cite{kim2023neuralfield,lin2023infinicity,chen2023scenedreamer}.
While the former approaches usually provide more details, they are limited to generating a fixed subset of possible scenes and require specialized data to train.
The latter approaches are theoretically more flexible but are currently bounded by their 3D representation power (hence producing fewer details), which is a problem we address in this work.

\begin{figure*}
    \centering
    \includegraphics[width=\textwidth]{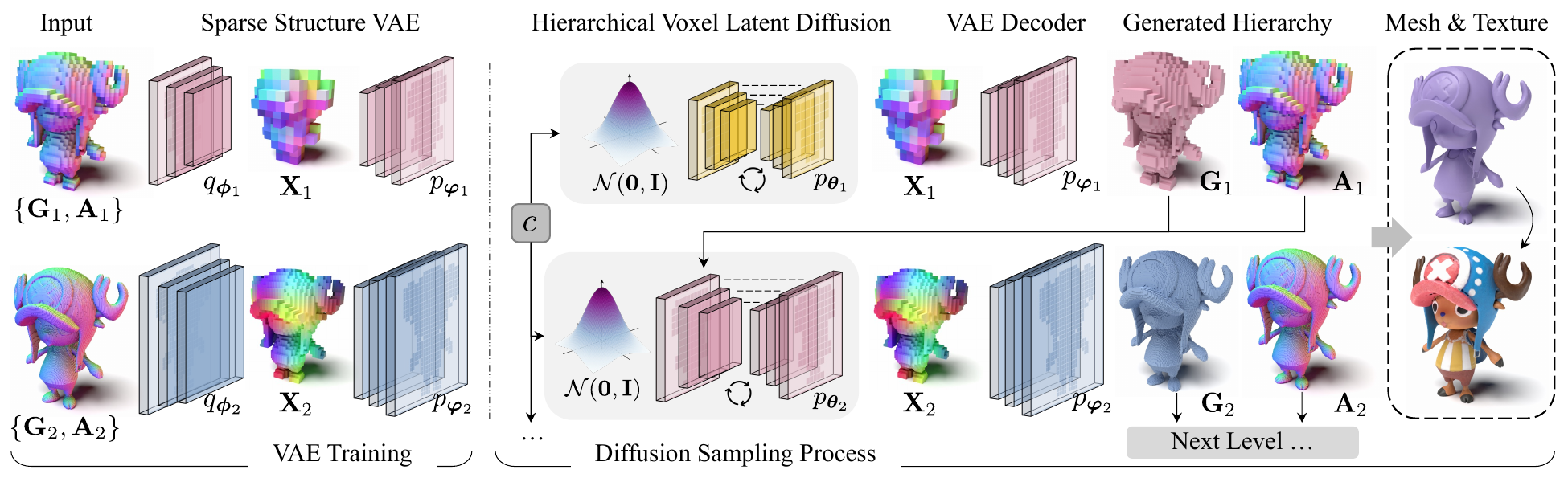}
    \vspace{-1.5em}
    \caption{\textbf{Method.} Sparse voxel grids within the hierarchy are first encoded into compact latent representations using a sparse structure VAE. The hierarchical latent diffusion model then learns to generate each level of the latent representation conditioned on the coarser level in a cascaded fashion. The generated high-resolution voxel grids contain various attributes for different applications.  Note that technically $\Latent_1$ is a dense latent grid, but illustrated as a sparse one for clarity.}
    \label{fig:architecture}
    \vspace{-1em}
\end{figure*}

\parahead{3D Representation for Generative Tasks}%
\emph{Point clouds}~\cite{achlioptas2018learning,luo2021diffusion,zeng2022lion,nakayama2023difffacto} are flexible and adaptive, but cannot represent solid surfaces and pose challenges in architecture design.
\emph{Triangle meshes}~\cite{gao2019sdmnet,nash2020polygen} are more expressive but are limited to a fixed topology and hence hard to optimize.
\emph{Neural fields}~\cite{luo2022surfgen,chen2019learning,nam20223dldm} encode scene geometry implicitly in the network weights and lack an explicit inductive bias for effective distribution modeling.
\emph{Tri-planes}~\cite{chen2022tensorf,gao2022get3d,NFD} can represent objects at high resolutions with reduced memory footprint, but fundamentally lack a geometric bias except for large axis-aligned planes, posing challenges when modelling larger scenes with complex geometry.
Comparably, \textit{voxel grids}~\cite{wu2017learning,NWD,sella2023vox} are flexible, expressive for both chunky and thin structures, and support fast querying and processing. 
\emph{Sparse} voxel hierarchies do not store voxel information for empty regions and hence are more efficient.
\change{A popular approach in the literature is to implement these using octree variants~\cite{tatarchenko2017octree,tang2021octfield,ibing2021octree,wang2022dual} or hash tables~\cite{tang2022torchsparse}.}
However, previous works either focus only on geometry~\cite{tatarchenko2017octree,zhang2021learning}, are limited to an effective resolution of  $256^3$~\cite{ibing2021octree}, do not consider hierarchical generation~\cite{Zheng_2023}, or are not evaluated on real-world datasets~\cite{li2023diffusion}.
In contrast, our method can generate high-resolution shapes from a hierarchical latent space, and is evaluated on large-scale, real-world scene datasets.

\section{Method}
\label{sec:method}

Our goal is to learn a generative model of large-scale 3D scenes represented as sparse voxel hierarchies. The hierarchy comprises of $L$ levels of coarse-to-fine voxel grids $\GridSet = \{ \Grid_1, \dots, \Grid_L \}$ and their associated per-voxel attributes $\AttrSet = \{ \Attr_1, \dots, \Attr_L \}$ such as normals and semantics.
Finer grids $\Grid_{l+1}$ with smaller voxel sizes are strictly contained within the coarser grids $\Grid_{l}$ for $l = 1, \dots, L-1$, and the finest level of grid $\Grid_L$ contains the maximum amount of details.

Our method trains a hierarchy of latent diffusion models over the sparse voxel grids $\GridSet$ encoded by a hierarchy of sparse structure VAEs, as summarized in \cref{fig:architecture}.
We first introduce the sparse structure variational autoencoder (VAE) that learns a compact latent representation of voxel grids in \cref{subsec:method:vae}.
Then we describe our full diffusion probabilistic model that learns the joint distribution of the latent representation and the sparse voxel hierarchy in \cref{subsec:method:dm}.
The training and sampling procedures are described in \cref{subsec:method:training}, followed by the implementation details in \cref{subsec:method:impl}.

\vspace{-2mm}
\subsection{Sparse Structure VAE}
\label{subsec:method:vae}
\vspace{-1mm}

\parahead{Motivation}
The sparse structure VAE is designed to learn a compact latent representation of each voxel grid within the hierarchy and its associated attributes.
Instead of directly modeling their joint distribution that comprises a mixture of continuous and discrete random variables, we encode them into a unified continuous latent representation, which is not only friendly to the downstream diffusion models during training and sampling~\cite{vahdat2021score,zeng2022lion}, but also facilitates the formulation of a hierarchical probabilistic model which we aim to demonstrate.
Additionally, the latent representation, encoded in a coarser spatial resolution, serves as a compact yet meaningful proxy that saves the computation while preserving the expressivity~\cite{rombach2022highresolution, gupta20233dgen}.
Represented as $\LatentSet = \{ \Latent_1, \dots, \Latent_L \}$, the latent is also a featurized sparse voxel hierarchy corresponding to a coarser version of $\GridSet$, with the voxel size of $\Latent_l$ being the same as $\Grid_{l-1}$.

\parahead{Network Design}
We choose to train separate VAEs that operate on each level $l$ of the hierarchy independently.
Hence for the ease of notation, we drop the subscript $l$ in the following discussion.
Here, we build the neural networks based on the operators over sparse voxel grids to model both the posterior distribution $q_\encw(\Latent | \Grid, \Attr)$ and the likelihood distribution $p_\decw(\Grid, \Attr | \Latent)$, with $\encw, \decw$ being the encoder and decoder weights respectively.

\begin{figure}
\vspace{-5.5mm}
    \centering
    \includegraphics[width=\linewidth]{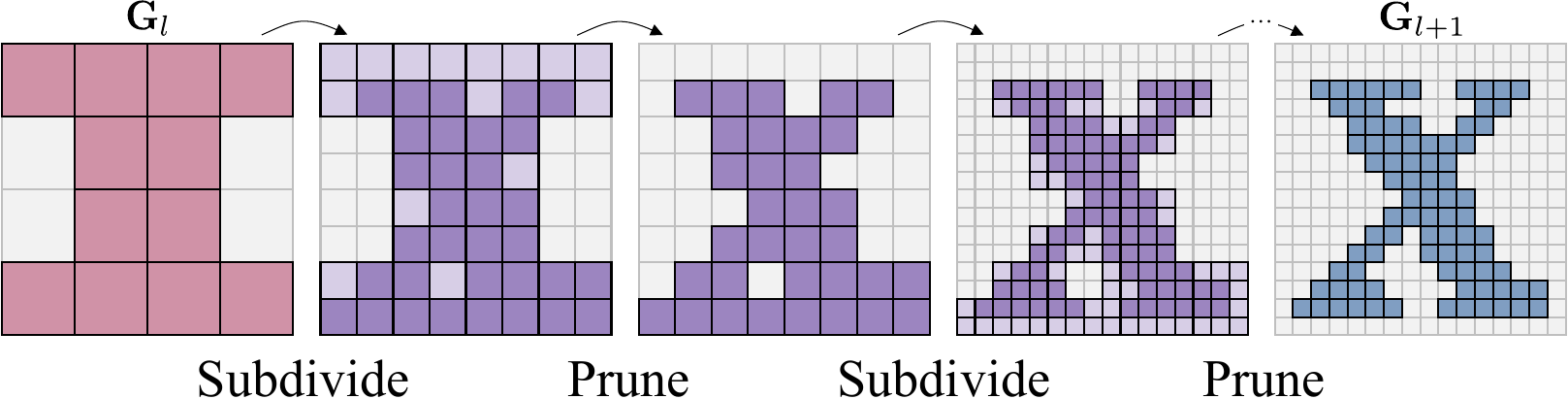}
    \vspace{-1.6em}
    \caption{\textbf{VAE Decoder Architecture.} Coarser levels of grids $\Grid_l$ are upsampled to finer grids $\Grid_{l+1}$ by iteratively subdividing existing voxels into octants and pruning excessive ones. Each \emph{level} may contain many upsampling  \emph{layers} that double the resolution.}
    \label{fig:decoder}
    \vspace{-1.2em}
\end{figure}

For the encoder, we utilize the sparse convolutional neural network to process the input $\Grid$ and $\Attr$ by alternatively applying sparse convolution and max pooling operations, downsampling to the resolution of $\Latent$.
For the decoder, we borrow the structure prediction backbone from \cite{huang2023neural} that allows us to predict novel sparse voxels that are not present in the input.
It starts from $\Latent$ and proceeds by progressively pruning excessive voxels and subdividing existing ones based on the prediction of a subdivision mask, and finally reaching the resolution of $\Grid$ after several upsampling layers.
An illustration of the above decoding scheme is shown in \cref{fig:decoder}.

\vspace{-1mm}
\subsection{Hierarchical Voxel Latent Diffusion}
\label{subsec:method:dm}

\parahead{Probabilistic Modeling}
Existing 3D generation literature~\cite{gupta20233dgen,nam20223dldm} typically uses one level of latent diffusion (\ie $L=1$).
While this is sufficient to generate intricate scenes containing one single object, the resolution is still far from enough to generate large-scale outdoor scenes.
The limited scalability of their underlying 3D representation and the absence of probabilistic modeling to capture the coarse-to-fine nature of the data hinder the effectiveness of these methods.
We solve this by marrying a hierarchical latent diffusion model~\cite{blattmann2023align, JMLR:v23:21-0635} with the sparse voxel representation.
Specifically, we propose the following factorization of the joint distribution of grids and latents:
\vspace{-1em}
\begin{equation}
    \label{eq:joint}
    p(\GridSet, \AttrSet, \LatentSet) = \prod_{l=1}^L p_{\decw_l}(\Grid_l, \Attr_l | \Latent_l) p_{\dmw_l}(\Latent_l | \Condition_{l-1}),
\end{equation}
where $\Condition_{l-1}$ is the condition from the coarser level, with:
\begin{equation}
    \Condition_{l} = \begin{cases}
        c, & l = 0\\
        \{\Grid_l, \Attr_l, c\}, & l > 0
      \end{cases},
\end{equation}
with $c$ being an optional global condition such as a category label or a text prompt, and $p_{\dmw_l}(\cdot)$ instantiated as a diffusion model with parameter $\dmw_l$ which we elaborate on later.

The above factorization assumes the Markov process (\ie level $l$ is only conditioned on its coarser level $l-1$), which is naturally induced from the geometric nature of the data.
By doing so, we reduce the layers in each level of VAE and amortize both the computation and the representation power across multiple levels (see \cref{subsec:exp:ablation} for empirical proof).
Additionally, such a factorized modeling endows us with utmost flexibility, enabling \emph{user controls} by editing or re-sampling grids from different levels.

\parahead{Diffusion Model $p_\dmw$}
Here we omit subscript $l$ again for clarity.
A diffusion stochastic process transforms a complicated distribution of a random variable into the unit Gaussian distribution $\Latent_0 \sim \Gauss(\Zero,\Identity)$ by iteratively adding white noise to it, following a Markov process~\cite{ho2020denoising}. One commonly used instantiation is the following:
\begin{equation}
    \Latent_t | \Latent_{t-1} \sim \Gauss(\sqrt{1-\beta_t} \Latent_{t-1}, \beta_t \Identity),
\end{equation}
where $0 < \beta_t \ll 1$ controls the amount of noise added for each step.
The reverse process, on the other hand, removes the noise iteratively and reaches data distribution $\Latent_T$ within a discrete number of steps $T$. It is usually modeled as:
\begin{equation}
    \Latent_{t-1} | \Latent_t \sim \Gauss(\netmu_\dmw (\Latent_t, t), \frac{1 - \bar{\alpha}_{t-1}}{1-\bar{\alpha}_t} \beta_t \Identity),
\end{equation}
with $\alpha_t = 1 - \beta_t$, $\bar{\alpha}_t = \prod_{s=0}^t \alpha_s$, and $\netmu_\dmw$ a parametrized learnable module.
In practice, we re-parametrize $\netmu_\dmw$ as:
\vspace{-2mm}
\begin{equation}
    \netmu_\dmw = \sqrt{\alpha_t} \Latent_t - \beta_t \sqrt{\frac{\bar{\alpha}_{t-1}}{1-\bar{\alpha}_t}} \netv_\dmw,
\end{equation}
so that the learnable module predicts $\netv$ instead. This is in accordance with the $\mathbf{v}$-parameterization in \cite{salimans2022progressive} that has been shown to facilitate training.

We instantiate $\netv_\dmw(\cdot)$ as a 3D \emph{sparse} variant of the backbone used in \cite{dhariwal2021diffusion}, ensuring the grid structure of the decoded output from $\netv_\dmw$ matches the input.
To inject condition $\Condition_{l-1}$, we directly concatenate the feature from $\Attr_{l-1}$ with the network input recalling that $\Latent_{l}$ also shares the same grid structure with $\Grid_{l-1}$.
Timestep condition is implemented using AdaGN~\cite{rombach2022highresolution} and textual condition $c$ is first CLIP-encoded~\cite{radford2021learning} and then injected using cross attention.

\begin{figure*}[t]
    \centering
    \includegraphics[width=\textwidth]{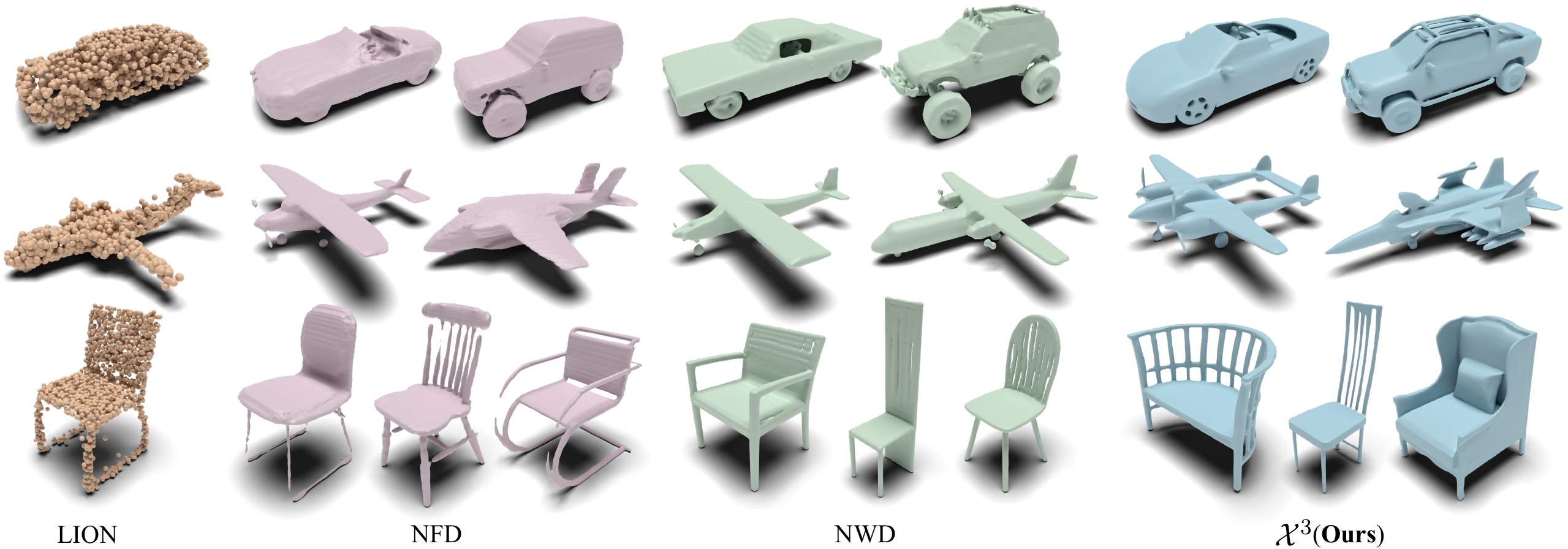}
    \vspace{-1.5em}
    \caption{\textbf{ShapeNet~\cite{Shapenet} Qualitative Comparison.} We show comparison of our method with LION~\cite{zeng2022lion}, NFD~\cite{NFD}, and NWD~\cite{NWD}. Our method is capable of generating intricate geometry and thin structures. Best viewed with 200\% zoom-in.}
    \label{fig:shapenet}
    \vspace{-1.5em}
\end{figure*}

\subsection{Training and Sampling}
\label{subsec:method:training}

\parahead{Loss Functions}
We train the VAE and the diffusion model level-by-level independently.
During the training of the level-$l$ VAE, we employ the following loss function:
\begin{equation}
\begin{aligned}
    \loss_{l}^{\text{VAE}} = & \; \expec_{\{\Grid_l, \Attr_l\}} [ \expec_{\Latent_l \sim q_\encw} [ \text{BCE}(\Grid_l, \tilde{\Grid}_l) + \\ & \loss_l^{\text{Attr}}(\Attr_l, \tilde{\Attr}_l) ] + \lambda \; \infdiv{q_{\encw}(\Latent_l)}{p(\Latent_l)} ],
\end{aligned}
\end{equation}
where $\tilde{\Grid}_l, \tilde{\Attr}_l$ is the output of the VAE decoder $\decw$ given $\Latent_l$, and $\loss_l^\text{Attr}$ is the loss supervising the attribute predictions (\eg, TSDF, semantics, etc.) with its specific form postponed to the supplementary material.
$\text{BCE}(\cdot)$ is the binary cross entropy on the grid structure,
making $p_\decw$ a mixed product distribution. 
$\infdiv{\cdot}{\cdot}$ is the KL divergence between the posterior and the prior $p(\Latent_l)$, which we set to unit Gaussian $\Gauss(\Zero, \Identity)$, and $\lambda$ is its weight.

The training loss for the diffusion model is:
\begin{equation}
    \loss_{l}^{\text{DM}} = \expec_{t,\Latent_l,\bm{\varepsilon}\sim \Gauss(\Zero, \Identity)} \left[ \lVert \netv_{\dmw_l} (\Latent_{l,t}, t) - \netv_\text{ref} \rVert^2_2 \right],
\end{equation}
where $\netv_\text{ref} = \sqrt{\bar{\alpha}_t} \bm{\varepsilon} - \sqrt{1-\bar{\alpha}_t} \Latent_l, t \sim [1, T]$, $\Latent_l$ is sampled from the VAE posterior, and $\Latent_{l,t} = \sqrt{\bar{\alpha}_t} \Latent_l + \sqrt{1-\bar{\alpha}_t} \bm{\varepsilon} $.

\parahead{Sampling}
To sample from the joint distribution of \cref{eq:joint}, one starts by drawing the coarest latent $\Latent_1$ from the diffusion model $p_{\dmw_1}$. Then, the decoder $p_{\decw_1}$ is used to generate the coarsest grid $\Grid_1$ and its associated attributes $\Attr_1$ (which is then optionally refined by the refinement network).
Conditioned on $\Condition_1 = \{\Grid_1, \Attr_1, c\}$, the diffusion model $p_{\dmw_2}$ is used to generate the next level of latent $\Latent_2$, and the process goes on until the highest resolution of $\{\Grid_L, \Attr_L\}$ is met.
We include TSDF in $\Attr_L$ for all our experiments, which enables us to decode high-resolution meshes.
For other tasks such as perception, we further allow for decoding other attributes such as semantics.
We use DDIM~\cite{DDIM} as our sampler.

\subsection{Implementation Details}
\label{subsec:method:impl}

In practice, we find the following implementation details, specially tuned for the sparse voxel hierarchy generation case, to be helpful for better results:
(1) \textbf{Early dilation}. In network layers with larger voxel sizes, we dilate the sparse voxel grids by one, so that the halo regions of the sparse topology also represent non-zero features. This helps later layers to better capture the local context and generate smooth structures.
(2) \textbf{Refinement network}. An inherent problem of our factorized modeling is error accumulation, where higher-resolution grids cannot easily fix the artifacts from prior layers. We mitigate this by appending a refinement network to the output of the VAE decoder that refines $\Grid_l$ and $\Attr_l$. The architecture of the refinement network is similar to \cite{huang2023neural}, and its training data is augmented by adding noise to the posterior of the VAE~\cite{JMLR:v23:21-0635} before being decoded.
Last but not least, our architecture and training details can be found in the supplementary.

\parahead{Sparse 3D Learning Framework} 
The use of sparse voxel grids motivates and enables us to build a custom 3D deep learning framework for sparse data in order to support higher spatial resolution and faster sampling.
To this end, we leverage the VDB~\cite{VDB} structure to store our sparse 3D voxel grid.
Thanks to its compact representation (taking only 11MB for 3.4 million of voxels) and fast look-up routine, we are able to implement common neural operators such as convolution and pooling in a very efficient manner.
Fully operating on the GPU (including grid building), our framework is able to process a 3D scene with $1024^3$ in milliseconds, runs $\sim$3$\times$ faster with $\sim$0.5$\times$ of the memory usage than the current state-of-the-art sparse 3D learning framework TorchSparse~\cite{tang2022torchsparse}.
All our architectures are based on this custom framework.

\vspace{-1mm}
\section{Experiments}
\label{sec:experiments}

\begin{table}[t]
\centering
\footnotesize
\setlength{\tabcolsep}{2mm}
\scalebox{0.88}{
\begin{tabular}{lcccccc}
\toprule
& \multicolumn{2}{c}{Airplane} & \multicolumn{2}{c}{Chair} & \multicolumn{2}{c}{Car} 			 \\ 
\cmidrule(lr){2-3} 
\cmidrule(lr){4-5} 
\cmidrule(lr){6-7}
    & CD & EMD & CD & EMD & CD & EMD 			 \\   
\midrule
\multicolumn{7}{l}{\textit{Point-based method}} \\
PVD~\cite{PVD} & 69.55 & 60.89 & 57.68 & 54.95 & 64.89 & 54.61 \\     
LION~\cite{zeng2022lion} & 65.10 & 60.15 & 56.72 & 54.28 & 60.61 & 54.94 \\   
\midrule
\multicolumn{7}{l}{\textit{Triplane-based method}} \\
NFD~\cite{NFD} & 57.55 & 53.47 & 54.87 & 54.06 & 69.49 & 71.96 \\ 
\midrule
\multicolumn{7}{l}{\textit{Dense voxel-based method}} \\
NWD~\cite{NWD} & 59.78 & 53.84 & 56.35 & 57.98 & 61.75 & 58.54 \\ 
\midrule
\change{LAS-Diffusion}~\cite{Zheng_2023} & 71.29 & 56.93 & 55.17  & 55.02 & 75.03 & 72.10 \\     
\change{3DShape2VecSet}~\cite{zhang20233dshape2vecset}& 62.75 & 61.01 & 54.06 & 56.79 & 86.85 & 80.91 \\ 
\midrule
\textbf{Ours} & \textbf{52.85} & \textbf{49.75} & \textbf{53.99} & \textbf{48.60} & \textbf{57.96} & \textbf{54.43} \\                                                
\bottomrule                                     
\end{tabular}
}
\vspace{-0.5em}
\caption{\textbf{1-NNA Comparison on ShapeNet~\cite{Shapenet}.} The lower the better. Best scores highlighted in bold.}
\label{table:shapenet}
\vspace{-1.5em}
\end{table}

We conduct comprehensive experiments to evaluate the performance of our model. First, we demonstrate \ShortName's ability to perform unconditional object-level 3D generation using ShapeNet~\cite{Shapenet} (\cref{subsec:exp:shapenet}), and conditional 3D generation from category and text using Objaverse~\cite{Objaverse}  (\cref{subsec:exp:objaverse}). Next, we showcase high-resolution outdoor scene-level 3D generation using both the Karton City~\cite{kartoncity} and Waymo~\cite{waymo} datasets (\cref{subsec:exp:scene}), which is one of the first results of this kind. Finally, we conduct ablation studies for our design choices (\cref{subsec:exp:ablation}). Please refer to the supplementary for further results.

\subsection{Object-level 3D Generation on ShapeNet}

\label{subsec:exp:shapenet}
\vspace{-1mm}
\parahead{Dataset} 
To benchmark \ShortName against prior methods, we use the widely-adopted ShapeNet~\cite{Shapenet} dataset. Following the experimental setup in~\cite{pointflow, zeng2022lion, luo2021diffusion, NFD}, we choose three specific categories: \textit{Airplane}, \textit{Car} and \textit{Chair}, containing 4145, 7496, 6778 shapes respectively. 
To build the ground-truth voxel hierarchy for training, we voxelize each mesh at a $512^3$ resolution and use the train/val/test split from \cite{ChoyXGCS16}.

\parahead{Evaluation}
To evaluate the geometric quality of our synthesized output, we follow previous work~\cite{zeng2022lion, NWD, pointflow} and use \emph{1-NNA} as our main metric (with both Chamfer distance (CD) and earth mover's distance (EMD)). 1-NNA provides a comprehensive measure of both quality and diversity by measuring the distributional similarity between the generated shapes and the validation set~\cite{pointflow, zeng2022lion}. Please refer to the supplementary for more details and evaluations.

\parahead{Baselines}  
We compare \ShortName to state-of-the-art 3D generative models that leverage various latent and shape representations: \textit{PVD}~\cite{PVD}, \textit{LION}~\cite{zeng2022lion}, \textit{NFD~\cite{NFD}}, \textit{NWD}~\cite{NWD}, \change{\textit{LAS-Diffusion}~\cite{Zheng_2023}, and \textit{3DShape2VecSet}~\cite{zhang20233dshape2vecset}}. PVD and LION employ point clouds as both latent and output-shape representations. 
\change{NFD, NWD, and 3DShape2VecSet all use mesh as the output shape representation, but with different latent representations, \ie, triplanes, dense voxel grids, and unstructured latent vectors, respectively.}
\change{Although LAS-Diffusion also uses sparse voxels similar to ours, it does not allow for generating arbitrary attributes and scale only to $128^3$ resolutions.}
In contrast, our method uses a sparse voxel hierarchy as a latent representation and outputs a sparse voxel grid, which can be readily converted to a mesh. 

\begin{figure}
    \centering
    \includegraphics[width=\linewidth]{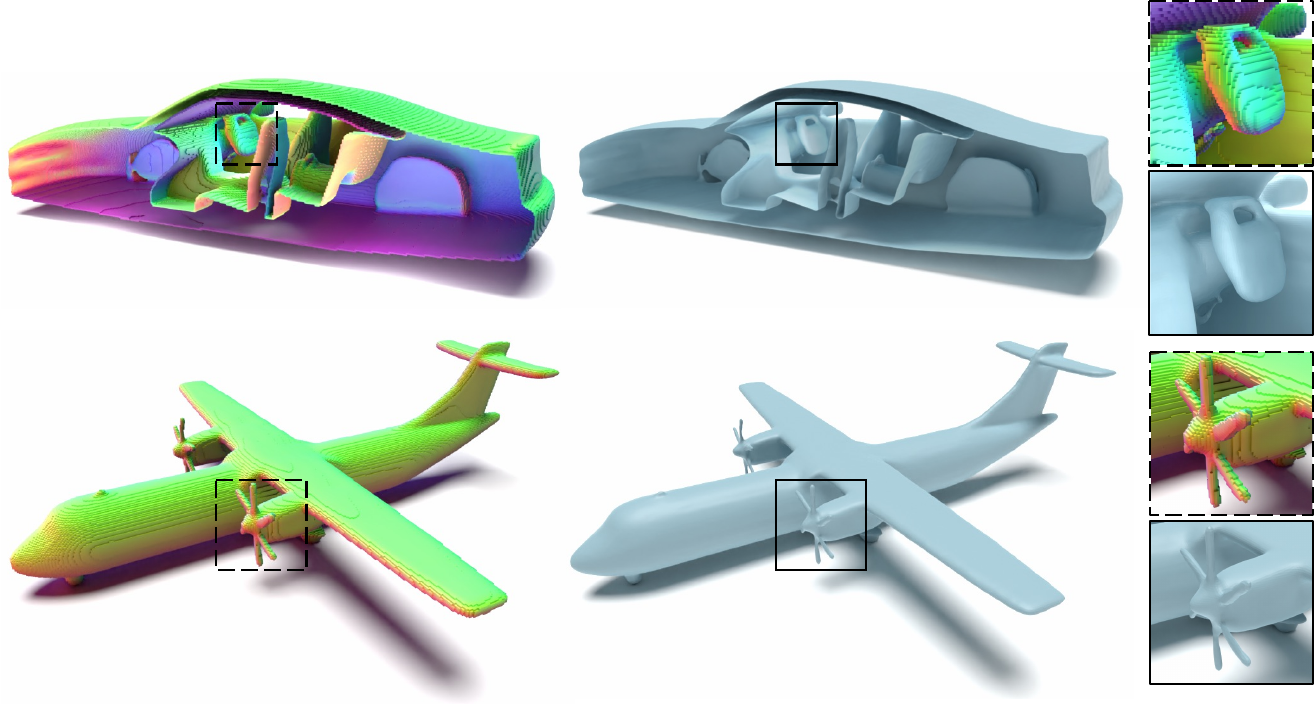}
    \vspace{-1.8em}
    \caption{\textbf{Close-up View of Our Generated Shape.} The voxel grid is colored by predicted normal. \ShortName is able to generate a high level of detail, such as the car interior and airplane propellers.}
    \label{fig:shapenet-detail}
    \vspace{-1.5em}
\end{figure}

\parahead{Results} 
\cref{table:shapenet} provides a quantitative comparison of \ShortName against baseline approaches and shows the superiority of our approach over past work. 
The point-based methods are naturally restricted to generating coarse shapes (\textit{i.e.}, 2048 points), while our method is able to generate millions of voxels ($500 \times$ larger). The triplane-based method (NFD) exhibits decent performance in categories such as Airplane and Chair with simple geometry. 
\change{However, its effectiveness diminishes for the Car category with intricate geometry (such as the suspension), underscoring the challenges to generate complex geometric structures using such representations.}
\change{3DShape2VecSet suffers from a similar issue where its latent representation compresses the information too aggressively.}
\change{While NWD and LAS-Diffusion are based on voxel grids, they are foundamentally limited by their low grid resolutions. The inverse wavelet transform in NWD is also lossy.}
In contrast, our method is able to generate high-resolution shapes with fine details, as shown in \cref{fig:shapenet-detail}.
\cref{fig:shapenet} shows a qualitative comparison, which is consistent with our quantitative results. 

\begin{figure}
    \centering
    \includegraphics[width=0.9\linewidth]{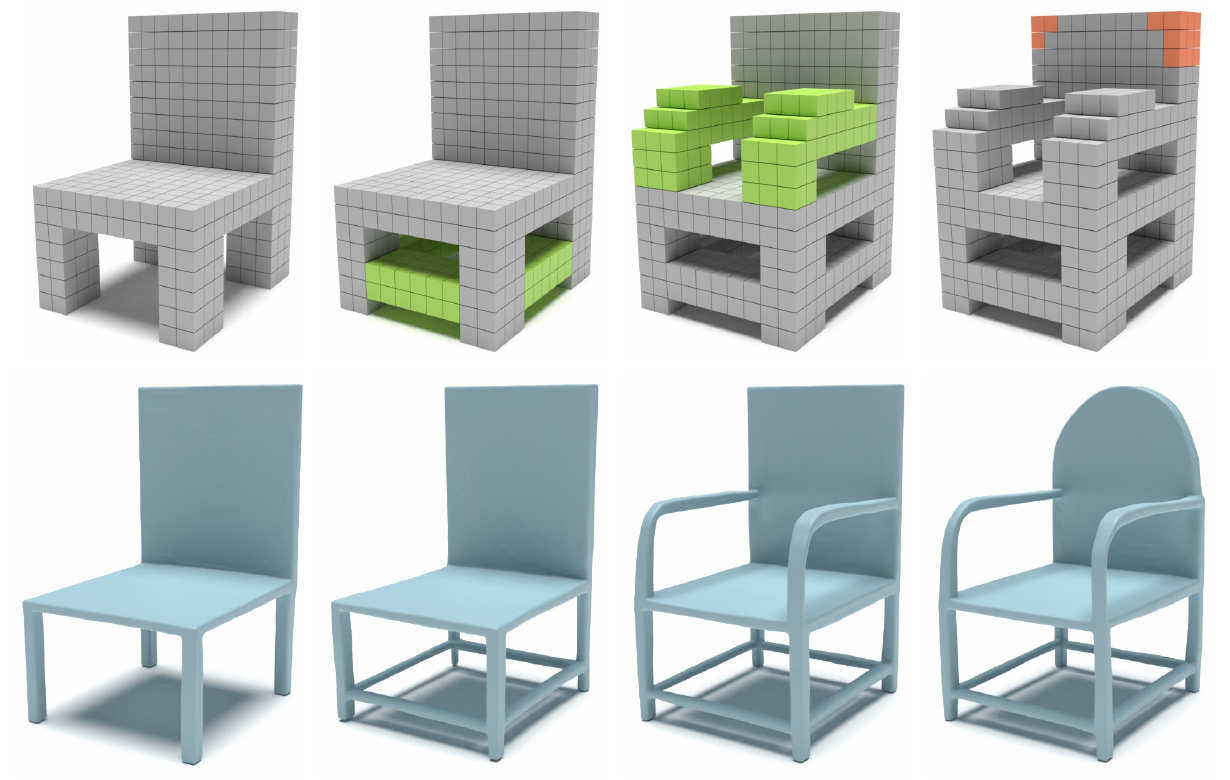}
    \caption{\textbf{User-guided Editing.} By adding (\textbf{\textcolor{mygreen}{green}}) and deleting (\textbf{\textcolor{myred}{red}}) coarse-level voxels, one can easily control the finer 3D shape.}
     \vspace{-1em}
    \label{fig:user-edit}
    \vspace{-0.5em}
\end{figure}

\parahead{User-guided Editing}
Our method is based on a sparse voxel hierarchy, which is a natural representation for user-guided editing. We demonstrate this ability by allowing users to edit the coarse-level shapes by adding or removing voxels in Goxel~\cite{Goxel}, a Minecraft-style 3D voxel editor. \cref{fig:user-edit} shows several examples of user-guided editing.

\subsection{Object-level 3D Generation on Objaverse}
\label{subsec:exp:objaverse}

\begin{figure*}
    \centering
    \includegraphics[width=\textwidth]{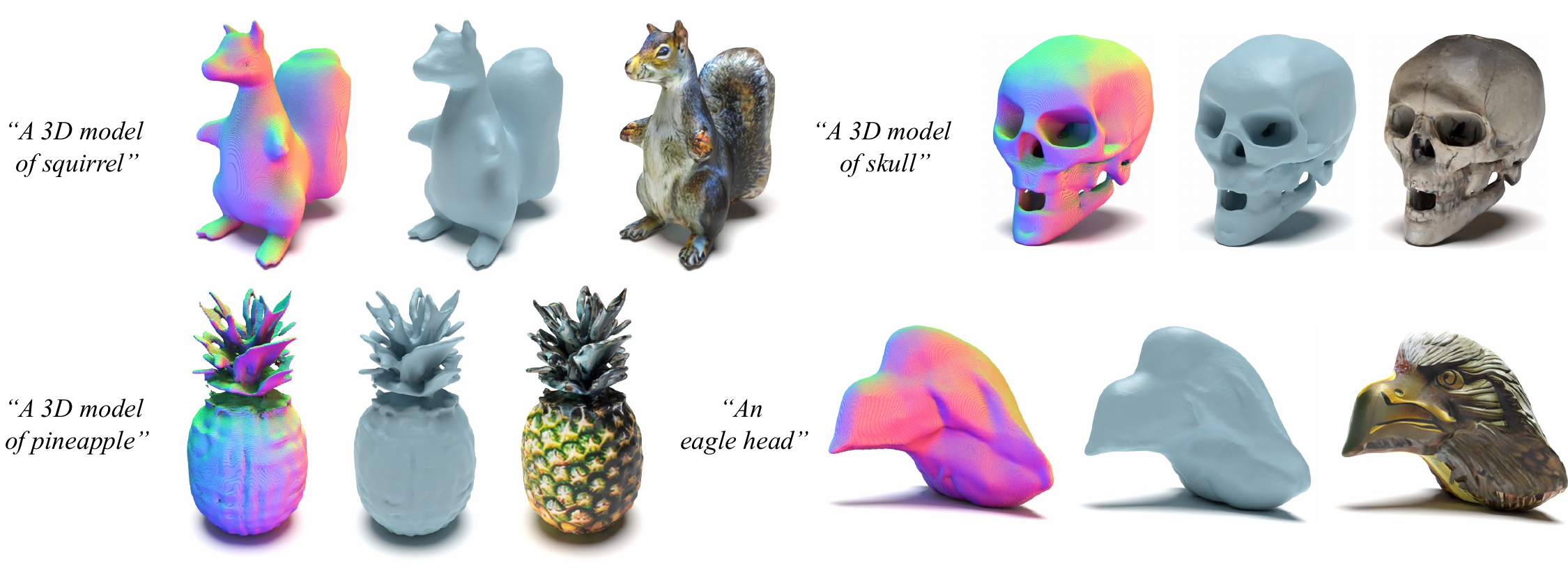}
    \vspace{-2.2em}
    \caption{\textbf{Text-to-3D Results on Objaverse~\cite{Objaverse}.} For each sample we show the input text prompt, the generated sparse voxel grid colored by normal, the extracted mesh, and the textured mesh (using \cite{texture}).}
    \label{fig:objaverse}
    \vspace{-2em}
\end{figure*}

\begin{figure}
    \centering
    \includegraphics[width=\linewidth]{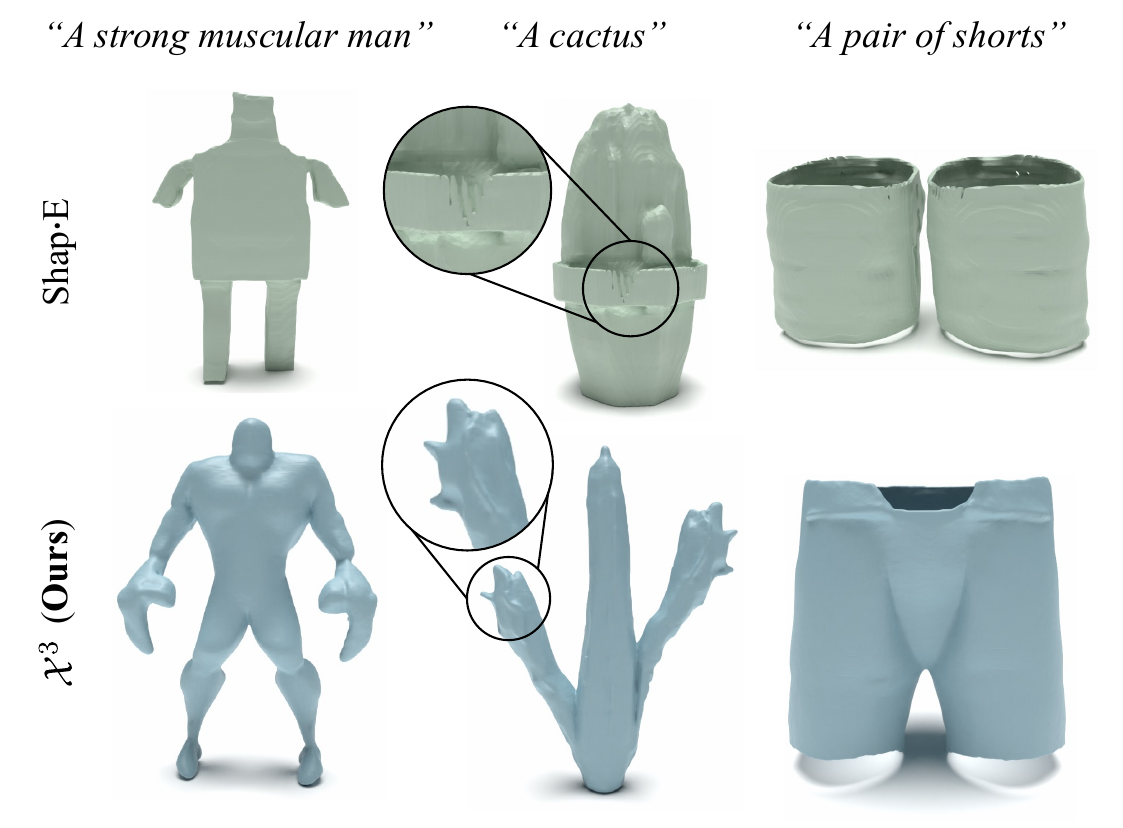}
    \vspace{-2em}
    \caption{\textbf{Comparison with Shap$\cdot$E~\cite{shape-e}}. We can generate high-quality shapes that better match the given prompts.}
    \label{fig:compare-shape}
    \vspace{-1.5em}
\end{figure}

\begin{figure}
    \centering
    \vspace{-1em}
    \includegraphics[width=0.95\linewidth]{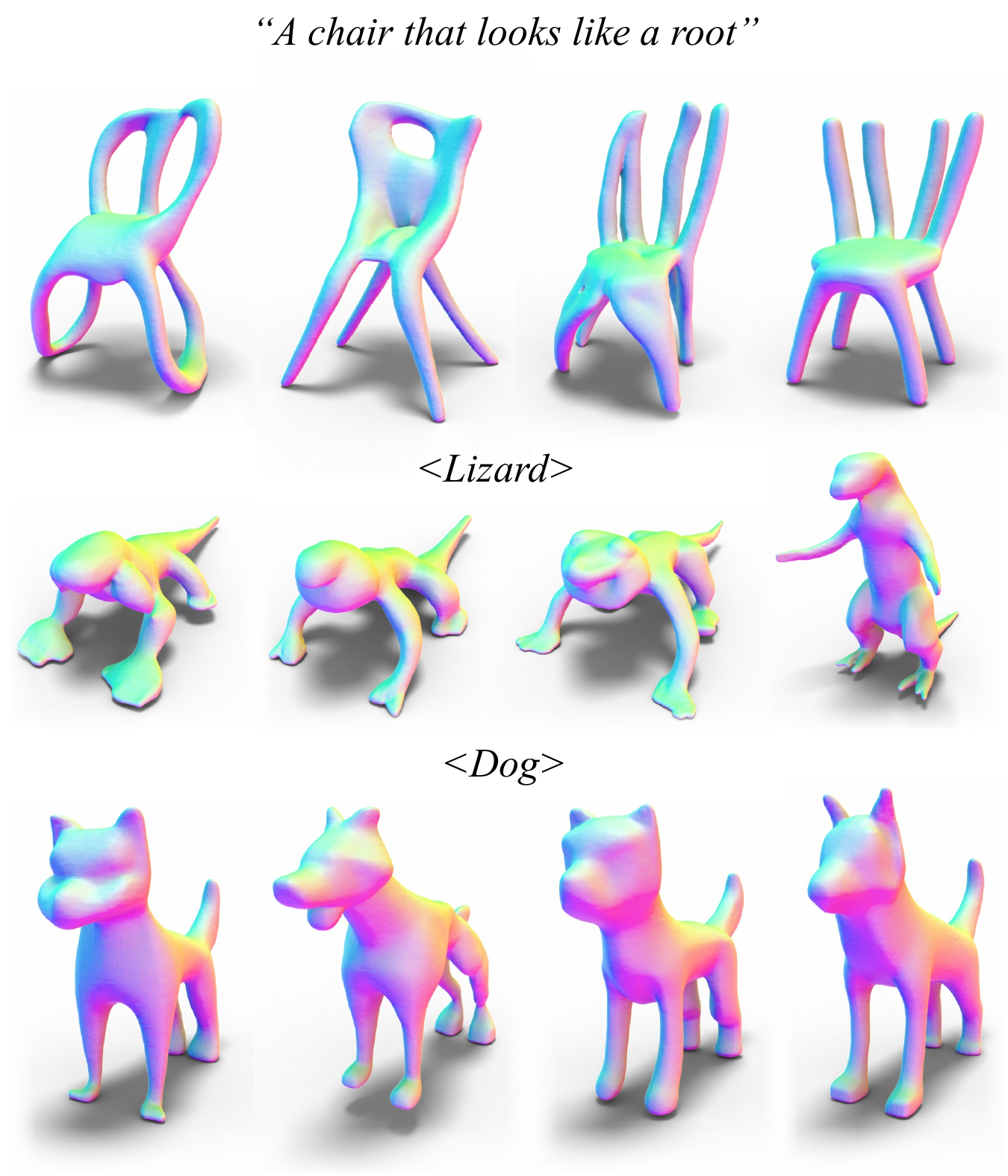}
     \vspace{-1em}
    \caption{\textbf{Diversity of Our Generated Shapes.} \ShortName can generate diverse shapes under the same text prompt or category label.}
    \vspace{-2em}
    \label{fig:shape-variation}
\end{figure}

\vspace{-1mm}
\parahead{Dataset}
To further demonstrate \ShortName's ability to perform object-level 3D generation, we evaluate it using the Objaverse~\cite{Objaverse} dataset, which offers approximately 800K publicly available 3D models. For the text-to-3D experiment, we use the descriptive text automatically generated by Cap3D~\cite{cap3d}. 
For category-conditional generation, we adopt the LVIS categorized object subset of \cite{Objaverse}, containing $\sim$40K 3D objects. 
We voxelize the 3D objects at a $512^3$ resolution to build the ground truth voxel hierarchy for training.

\parahead{Evaluation}
We conduct a user study on Amazon Mturk to compare our method with Shap$\cdot$E~\cite{shape-e}, a state-of-the-art text-to-3D method. Specifically, we gather $30$ prompts from popular text-to-3D works~\cite{shape-e, dreamgaussian, ntavelis2023autodecoding, point-e}.  For each prompt, we ask $30$ users to pick the 3D object (ours v.s. Shap$\cdot$E w/o texture) that better matches the text prompt and exhibits higher geometric fidelity. In total, we collect $900$ pairwise comparisons. In $79.2\%$ of the comparisons, participants vote for our generated 3D objects.

\parahead{Results}
We provide qualitative results for text-to-3D in \cref{fig:objaverse}, and category-conditional generation in \cref{fig:shape-variation}. The texture of our results is generated by off-the-shelf texture synthesis methods~\cite{texture,cao2023texfusion}. We also compare our method with Shap$\cdot$E~\cite{shape-e} in \cref{fig:compare-shape}. Our whole pipeline including generating the geometry and the texture for one object takes about 1 minute (30s for objects and 30s for textures). We observe that our method is able to generate more diverse 3D objects with higher geometric fidelity and finer details than Shap$\cdot$E, as shown in \cref{fig:shape-variation}.

\subsection{Large-scale Scene-level 3D Generation}

\label{subsec:exp:scene}
\vspace{-1mm}
\parahead{Dataset}
To demonstrate our model's scalability and ability to generate large-scale high-resolution scenes, we train and evaluate it on the Waymo Open Dataset~\cite{waymo} which contains $1000$ LiDAR scan sequences capturing different driving scenarios. Here, we extract the ground-truth dense scene geometry by accumulating the LiDAR scans and propagating the semantic labels.
To construct the ground-truth sparse voxel hierarchy, we crop each of the extracted scenes to \SI{102.4}{\meter}$\times$ \SI{102.4}{\meter} chunks and voxelize the points and meshes at a resolution of $1024^3$, resulting in a voxel size of \SI{10}{\centi\meter}.
To demonstrate the superiority of our representation power, we train a model on Karton City~\cite{kartoncity}, a custom synthetic dataset of 20 blocks of synthetic city scenes using the same resolution settings as Waymo. We divide the 20 blocks into train/val splits and randomly crop \SI{102.4}{\meter}$\times$ \SI{102.4}{\meter} chunks in each split to generate 900/100 unique train/val scenes.
\cref{fig:kartoncity} shows examples from the Karton City model and we include more results in the supplementary.

\begin{figure*}
    \centering
    \includegraphics[width=\textwidth]{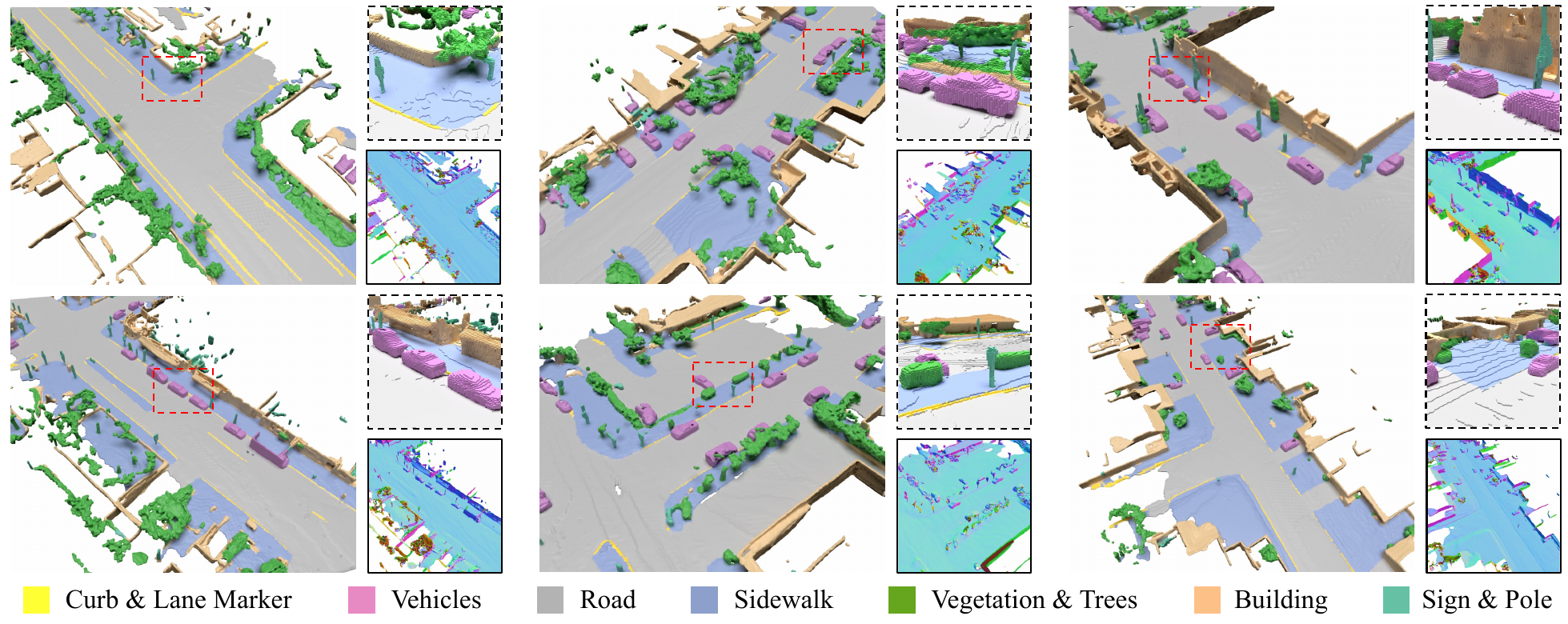}
     \vspace{-1.6em}
    \caption{\textbf{Unconditional Samples on the Waymo Open Dataset~\cite{waymo}.} The dashed boxes show a zoomed-in view and the solid boxes show the normal map for the extracted mesh. Best viewed with 200\% zoom-in.}
    \label{fig:waymo-uncond}
    \vspace{-1.5em}
\end{figure*}

\begin{figure}
    \centering
    \includegraphics[width=0.95\linewidth]{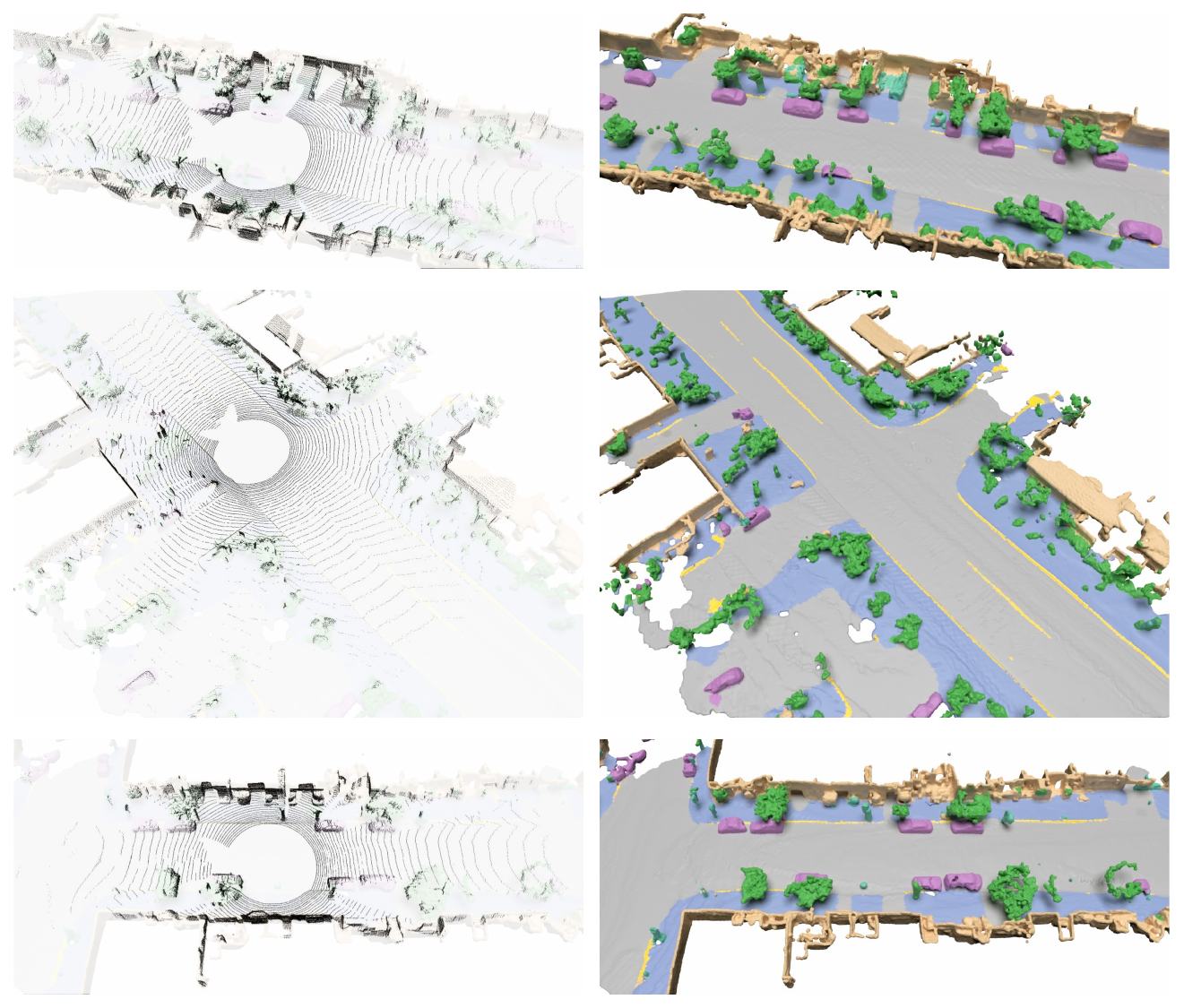}
    \caption{\textbf{Single-scan-conditioned Generation.} The left column shows the input LiDAR scan and the right column shows our generated semantic mesh conditioned on the input.}
    \label{fig:waymo-single-scan}
    \vspace{-1.5em}
\end{figure}

\parahead{Evaluation}
To evaluate the quality of our generated results, we perform a user study using Amazon Mturk. Here we show 30 users 30 pairs of scenes (totaling 900 pairwise comparisons) where one scene is sampled from the validation set and the other is a random output generated by our model. We ask each user to rank which scene is more realistic. Out of the 900 comparisons, $66.3\%$ favors our results over the ground truth, demonstrating that our generated outputs are of high quality. 
\cref{fig:waymo-uncond} shows several unconditional generations from our model as well as decoded semantics and normals.

\parahead{Single-Scan Conditioning}
We demonstrate that our model can be used to perform conditional generation on large-scale scenes. In this qualitative experiment, we condition the model on a single input LiDAR scan and generate a complete scene with normals and semantics. \cref{fig:waymo-single-scan} shows several completions using our method. 
Note that our input does not contain semantics, yet our model is able to generate plausible geometric and semantic completion results.
The supplementary shows additional details and figures.

\subsection{Ablation Study}
\label{subsec:exp:ablation}

\vspace{-1mm}
\parahead{Progressive Pruning}
We replace the progressive pruning part of our pipeline with a single pruning step for the $16^3 \rightarrow 128^3$ VAE on ShapeNet Chairs. We observe that the reconstruction accuracy (grid IoU) drops from $92.88 \% $ to 
$89.68 \%$, indicating that progressive pruning is critical for preserving shape details and injecting 3D inductive bias. Furthermore, the GPU memory usage also increases by a factor of $3\times$ when removing progressive pruning.

\parahead{Hierarchy Configuration}
As shown in \cref{table:ablation}, we compare the performance of our model with different hierarchy configurations on ShapeNet Chairs.  
We observe that: 
(1) the hierarchical model outperforms the single-level model, emphasizing the importance of a sparse voxel hierarchy in 3D generative modeling. 
(2) the model's performance is robust to the resolution of the initial hierarchy level. We find $16^3$ is sufficient for capturing the overall shape of the object.
(3) using two-level and three-level models achieve comparable performance. For unconditional generation, we use a two-level model for fast sampling. And for the user-editing setting, we use a three-level model for easier editing.

\begin{table}[t]
    \vspace{-0.8em}
\begin{center}
\footnotesize
\begin{tabular}{lcc}
\toprule
\multicolumn{1}{l}{Model} & CD (\%) & EMD (\%)  \\
\midrule
$16^3$ $\rightarrow$ $512^3$ & 59.31 & 57.46\\
$16^3$ $\rightarrow$ $128^3$ $\rightarrow$ $512^3$ & 53.99 & \textbf{48.60}  \\
$32^3$ $\rightarrow$ $128^3$ $\rightarrow$ $512^3$ & 55.39 & 51.40  \\
$4^3$ $\rightarrow$ $16^3$ $\rightarrow$ $128^3$ $\rightarrow$ $512^3$ & \textbf{52.88} & 53.62 \\
\bottomrule
\end{tabular}
\end{center}
\vspace{-1.2em}
\caption{ \textbf{Ablation of Different Resolutions and Depths of the Hierarchy.} Metrics are in 1-NNA. The lower the better.}
\label{table:ablation}
\vspace{-1em}
\end{table}

\begin{figure}
    \centering
    \includegraphics[width=\linewidth]{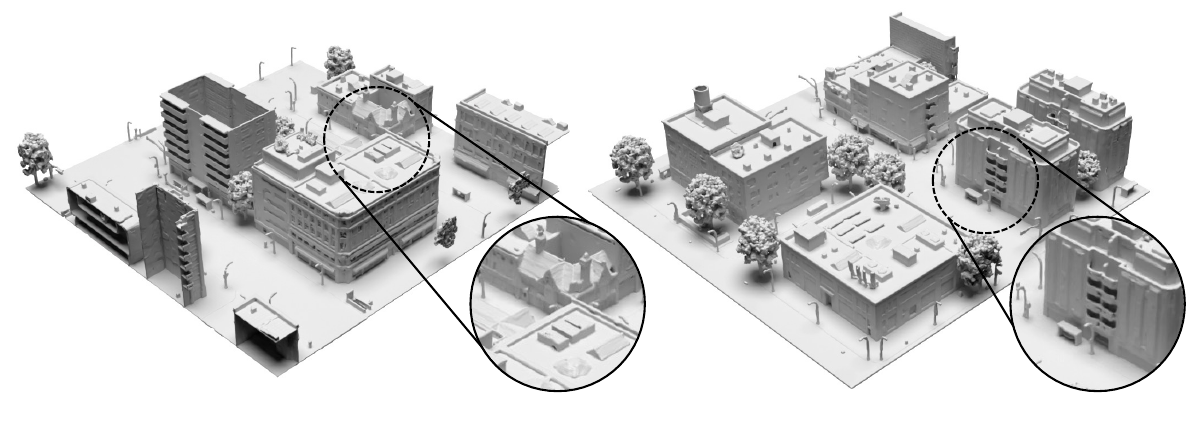}
     \vspace{-1.8em}
    \caption{\textbf{Unconditional Samples on Karton City~\cite{kartoncity}.}}
    \label{fig:kartoncity}
    \vspace{-1.8em}
\end{figure}

\vspace{-2mm}
\section{Discussion}
\label{sec:conclusion}
\vspace{-2mm}
\parahead{Conclusion}
We presented \LongName, a novel generative model for large-scale 3D scenes represented as a hierarchy of sparse 3D voxel grids.
We proposed a hierarchical voxel latent diffusion model that learns the joint distribution of the latent representation and the sparse voxel hierarchy.
The effectiveness of our method was demonstrated on both object-level and scene-level generation, reflecting our method's capability of generating high-resolution 3D scenes with fine details.

\parahead{Limitations and Future Work}
Due to current 3D datasets being still not on par with image datasets (such as \cite{LAION})
, our text-to-3D model is hard to deal with complex prompts.
In the future, we plan to extend our method in the setting of image-conditioning, as well as leveraging the learned prior as a fundamental model to support more downstream tasks.

\clearpage

{
    \small
    \bibliographystyle{ieeenat_fullname}
    \bibliography{main}
}

\clearpage
\onecolumn
\appendix
\appendixpage
In this supplementary material, we provide additional details on our method and experiments.
In \cref{sec:framework}, we describe our sparse 3D deep learning framework, and compare it to state-of-the-art implementations.
In \cref{sec:implementation}, we provide more implementation details for our method as well as precise definitions of our loss function and evaluation metrics.
In \cref{sec:more-results}, we provide more qualitative results on all the datasets we trained/evaluated on in the main paper.
We additionally provide a \textbf{supplementary video} in the accompanying files to better illustrate our results.

\section{Sparse 3D Learning Framework}
\label{sec:framework}

All of our networks are implemented using a customized sparse 3D deep learning framework built on top of PyTorch. 
To represent sparse grids of features and perform efficient deep-learning operations (convolution, pooling, etc.) over them, we leverage NanoVDB~\cite{nanovdb}, a GPU-friendly implementation of the VDB data structure~\cite{VDB}. 
A VDB tree is a variant of B+-Tree with four layers where the top layer is a hash table, followed by two internal layers (with branching factor $32^3, 16^3$), followed by leaf nodes with $8^3$ voxels. 

To demonstrate the effectiveness of our VDB-based deep learning framework, we benchmark it against TorchSparse~\cite{tang2022torchsparse}, a state-of-the-art sparse deep learning framework.
As shown in \cref{tab:benchmark}, our custom framework is both fast and memory-efficient, especially for large input grid resolutions.
Built upon the highly efficient VDB data structure, our 3D representation is compactly stored in memory and supports more efficient nearest neighbor lookup and processing than its hash table counterpart in \cite{tang2022torchsparse}.
Such a framework lays the foundation for our high-resolution 3D generative model and has potential to applications in many other downstream tasks such as reconstruction and perception.

\begin{table}[!htbp]
\centering
\footnotesize
\begin{tabular}{@{}lccccc@{}}
\toprule
                & \multicolumn{2}{c}{Voxel Grid Memory (MB) $\downarrow$} & \multicolumn{3}{c}{Convolution Forward Time (ms) $\downarrow$} \\
Grid Resolution & $512^3$  & $1024^3$   & $32^3$  & $256^3$  & $1024^3$ \\ 
\midrule
TorchSparse~\cite{tang2022torchsparse}     &   15.0           &   104.6         &    2.1               &  8.5       &   446.0      \\
\textbf{Ours}            &     \textbf{3.6}        &    \textbf{8.4}        &        \textbf{0.5}         &     \textbf{5.0}     &    \textbf{149.6}    \\ \bottomrule
\end{tabular}
\caption{Sparse 3D benchmark results.}
\label{tab:benchmark}
\end{table}

\section{Implementation Details}
\label{sec:implementation}

\subsection{Loss Definition}
Our model is able to output various attributes $\Attr$ defined on the voxel grids. 
Here we omit the level subscript $l$ for simplicity.
The direct output of the network at each voxel at each level includes surface normal $\bm{n} \in \RR^3$, semantic label $\bm{s} \in \RR^S$, and neural kernel features $\bm{\phi} \in \RR^4$.
Here the neural kernel features $\bm{\phi}$ are used for computing continuous TSDF values in 3D space for highly-detailed \emph{subvoxel}-level surface extraction (using the techniques from \cite{huang2023neural}), and it could also be replaced with implicit features $\bm{q}$ to extract TUDF values for open surfaces as in \cite{peng2020convolutional}.
The attribute loss $\loss^\text{Attr}$, as mentioned in Eq.~({\textcolor{blue}{6}}) of the main text, is a mixture of different supervisions, written as follows:
\begin{equation}
    \loss^\text{Attr} = \lambda_1 \underbrace{\lVert \bm{n} - \bm{n}_\text{GT} \rVert_2^2}_{\text{normal loss}} + \lambda_2 \underbrace{\text{BCE}(\bm{s}, \bm{s}_\text{GT})}_\text{semantic loss} + \lambda_3 \underbrace{\mathbb{E}_{\bm{x} \in \RR^3} \lVert f(\bm{x}) - \text{TSDF}(\bm{x}, \bm{X}_\text{GT}) \rVert_1 }_\text{surface loss},
\end{equation}
where $\bm{n}_\text{GT}$ and $\bm{s}_\text{GT}$ are the ground-truth normal and semantic label at each voxel, and $\bm{X}_\text{GT}$ is the ground-truth dense point cloud of the surface.
The surface loss is computed by sampling points $\bm{x}$ in the 3D space and comparing the predicted TSDF values $f(\bm{x})$ with the ground-truth TSDF values $\text{TSDF}(\bm{x}, \bm{X}_\text{GT})$.
To compute $f(\bm{x})$ given arbitrary input positions, we leverage the predicted neural kernels $\bm{\phi}$ to solve for a surface fitting problem as in \cite{huang2023neural}:
\begin{equation}
    f(\bm{x}) = \sum_v \alpha_v K(\bm{x}, \bm{x}_v) = \sum_v \alpha_v \bm{\phi}_v^\top \bm{\phi}(\bm{x}) K_b(\bm{x}, \bm{x}_v),
\end{equation}
where $v$ is the index of the voxels, $\bm{\phi}(\bm{x})$ is the neural kernel evaluated at the input position $\bm{x}$ using bezier interpolation from its nearby voxels, and $K_b(\bm{x}, \bm{x}_v) = B(\bm{x} - \bm{x}_v)$ is a shift-invariant Bezier kernel.
The coefficients $\alpha_v$ are obtained by performing a linear solve as detailed in \cite{huang2023neural}.
Similarly, for open surfaces we can replace the neural kernels $\bm{\phi}$ with implicit features $\bm{q}$ and define $f(\bm{x})$ as a local MLP function digesting trilinearly interpolated $\bm{q}$ at position $\bm{x}$~\cite{peng2020convolutional}.
We set $\lambda_1=1$, $\lambda_2=15$, and $\lambda_3=1$ in our experiment. 
For the KL divergence, we normalize it by the number of voxels of the voxel grid and then use a loss weight $\lambda=0.0015$ for all our experiments.

\subsection{Conditioning}
We explore diverse condition settings for our voxel diffusion models: 
(1) For the associated attributes from the previous level, we optionally concatenate them to the latent feature $\mathbf{X}$ before the latent diffusion. For example, for user-control cases, we do not concatenate them for flexibly adding or deleting voxels. (2) For the text prompts, we use cross-attention to fuse them into the latent. (3) For the category condition, we use AdaGN and fuse them with timestep embedding by adding. (4) For single scan conditions, we use an additional point encoder to quantize the single scan point cloud to a voxel grid and concatenate it with the latent feature $\mathbf{X}$.

\parahead{Micro-conditioning}
We found that the Waymo dataset suffers from missing voxels due to the sparsity of the LiDAR scans. To mitigate this issue, we use a micro-conditioning scheme following SD-XL~\cite{podell2023sdxl} to inject additional condition to the diffusion backbone describing the number of the voxels. This helps when the dataset itself contains multi-modal distributions, and allows fine-grained control of the generated scale of the scene.

\begin{figure*}
\centering
\begin{tabular}{ccc}
\includegraphics[width=0.32\linewidth]{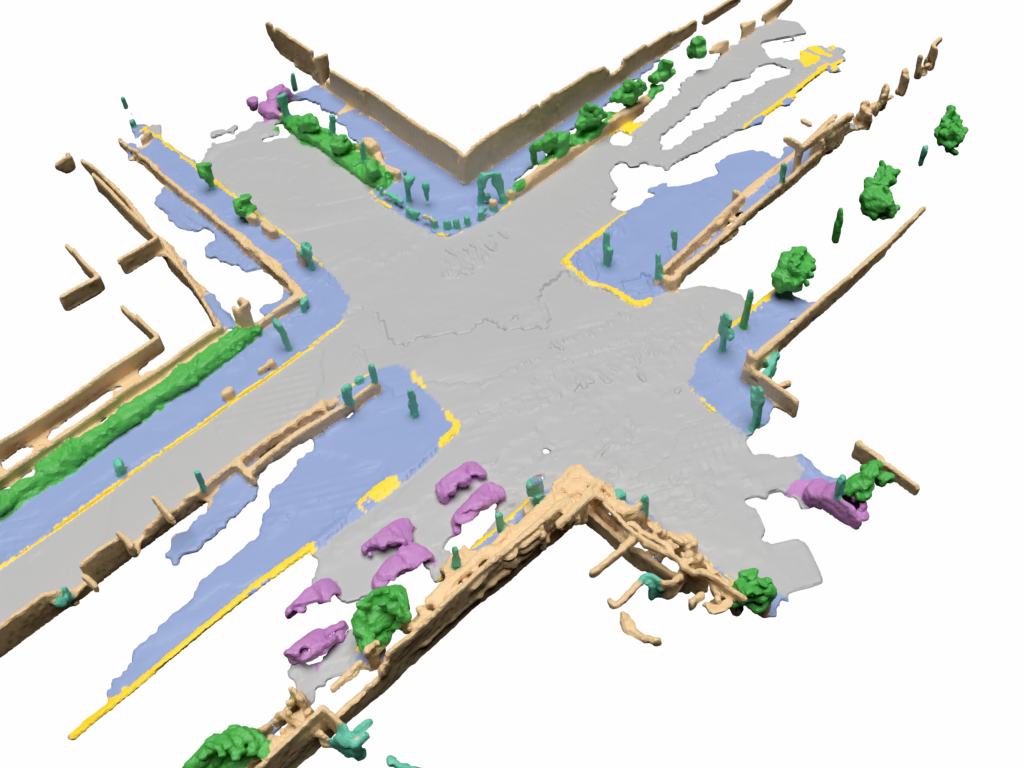}&
\includegraphics[width=0.32\linewidth]{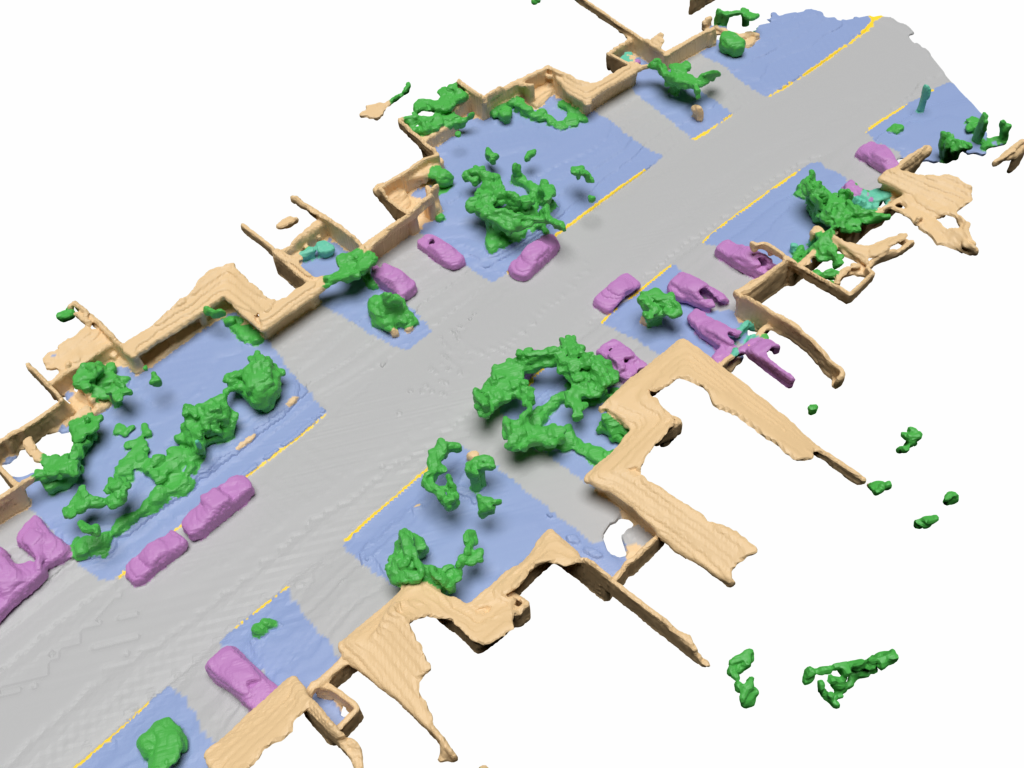}&
\includegraphics[width=0.32\linewidth]{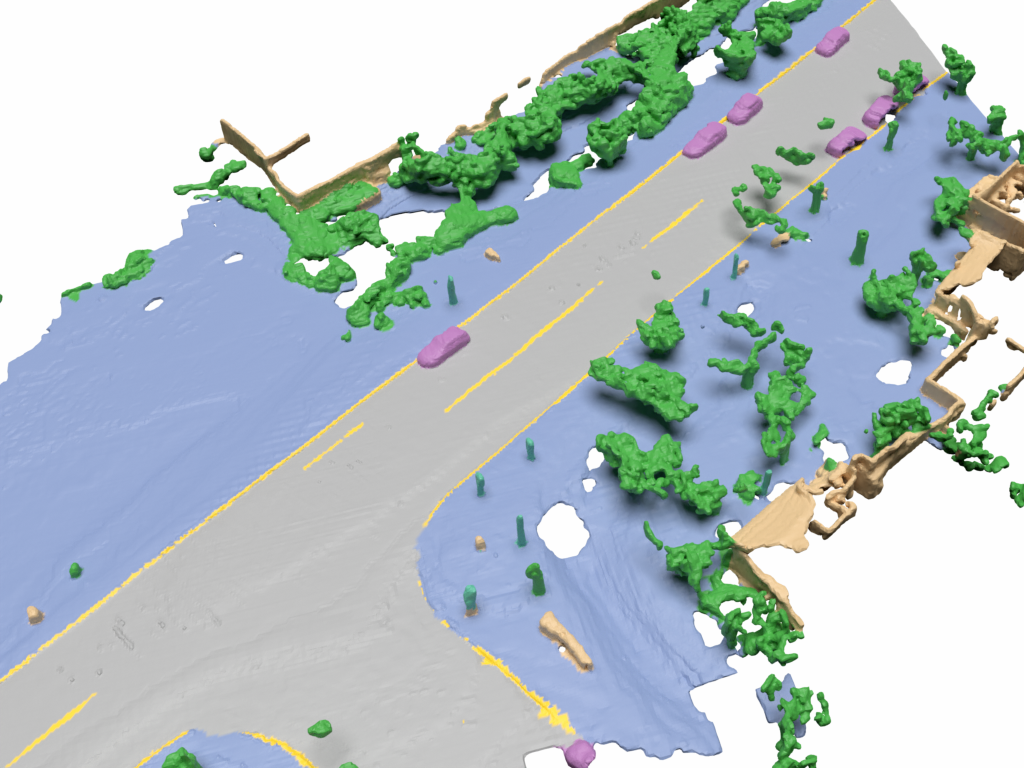}\\
\end{tabular}
\caption{Results of micro-conditioning on Waymo dataset. The voxel number conditioning increases from left to right. There is a clear trend of increasing number of voxels and more diverse contents in the sampled scenes.}
\label{fig:micro-conditioning}
\end{figure*}

\subsection{Texture Synthesis}
\begin{figure*}
    \centering
    \includegraphics[width=\linewidth]{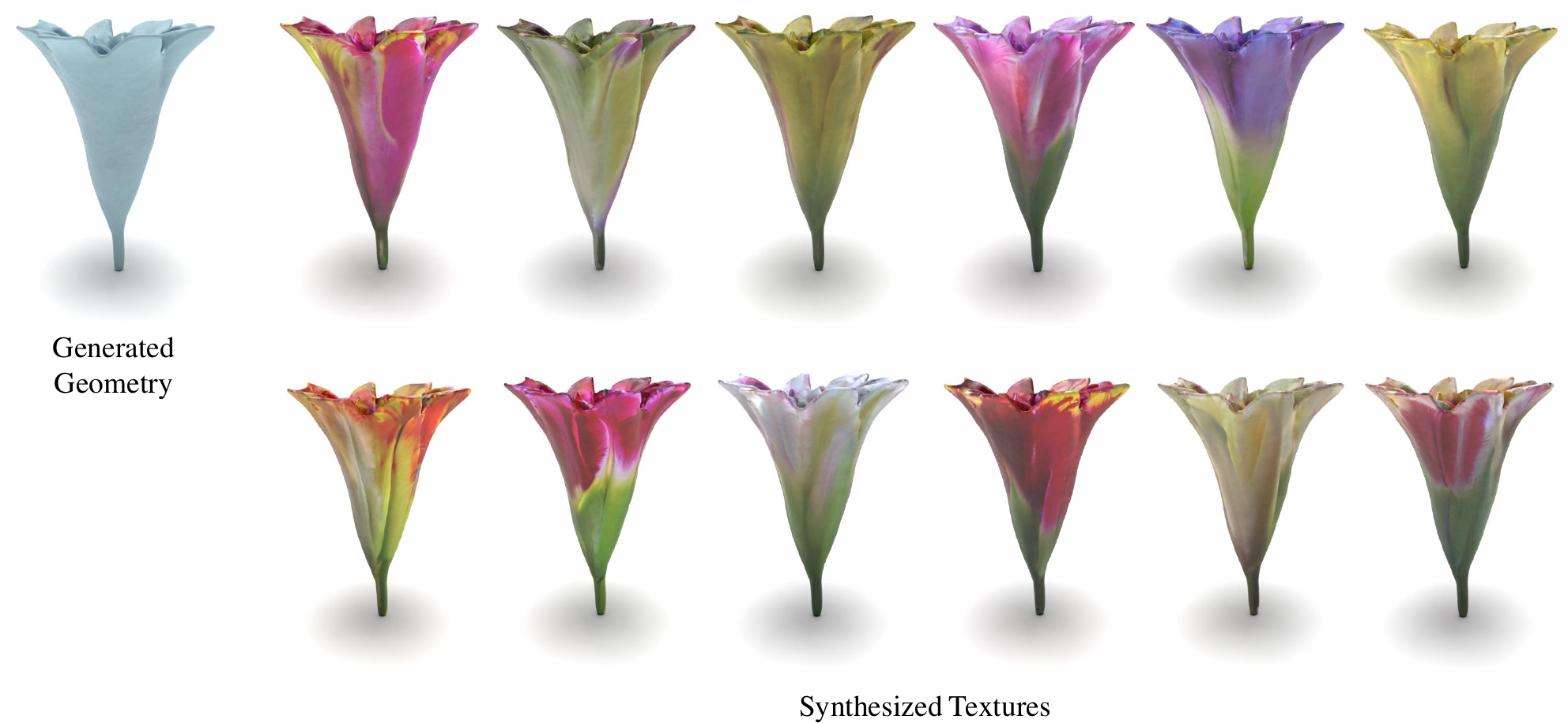}
    \caption{Diverse texture synthesis results. Based on the same generated geomtry, we could generate diverse textures by using TEXTrue~\cite{texture}.}
    \label{fig:objaverse-supplement-diversity}
\end{figure*}

While our model is focused on generating 3D geometry, we also explore the possibility of generating textures for the generated shapes.
To this end, we use a state-of-the-art texture generator TEXTure~\cite{texture} to create texture maps for the generated shapes.
The method works by applying a sequence of depth-conditioned stable diffusion models to multiple views of the shape.
Later steps in the process are conditioned on the previous steps, allowing the model to generate consistent textures with smooth transitions.
We choose to decouple the geometry and texture generation processes to allow for more flexibility and controlability -- \eg, given the same geometry, different textures can be generated and selected.
We demonstrate the effectiveness of the full pipeline on Objaverse dataset and use the same text prompts for both the geometry and the texture. Results are shown in \cref{fig:objaverse-supplement-diversity}.

\subsection{Network Architecture}
\parahead{Variational Autoencoders (VAE)}
We use a custom Autoencoder architecture for our VAE. Given an input voxel grid, $\Grid_l$ at level $l$, and associated per-voxel attributes $\Attr_l$, we first positionally encode each voxel using the same  function as~\cite{nerf} and then concatenate the positional encoding of each voxel with the corresponding attribute. We then apply a linear layer to the concatenated positional embedding and attribute to lift it to a $d$-dimensional feature (Where $d$ is chosen depending on the task and described in Table~\ref{table:train_vae}). Our VAE then applies successive convolution and max pooling layers, coarsening the voxels to a bottleneck dimension. When $l = 1$ (\textit{i.e.} the coarsest level of the hierarchy), we zero pad the bottleneck layer into a dense tensor, otherwise, the bottleneck is a sparse tensor. We then apply $4$ convolutional layers to convert the bottleneck tensor into a latent tensor $\Latent$ of the same shape and sparsity pattern as the bottleneck. Latent diffusion is done over the tensor $\Latent$. At the end of the decoder, we apply attributes-specific heads (MLPs) to predict the associated attributes within each voxel. Hyperparameters for our VAEs are listed in \cref{table:train_vae}.

\parahead{Diffusion UNet}
As mention in the main paper, we adopt a a 3D sparse variant of the backbone used in \cite{dhariwal2021diffusion} for our voxel latent diffusion. Hyperparameters for training them are in \cref{table:train}

\subsection{Training Details}
We train all of our models using Adam~\cite{adam} with $\beta_1 = 0.9$ and  $\beta_1 = 0.999$. We use an EMA rate of $0.9999$ for all experiments and use PyTorch Lightning~\cite{Falcon_PyTorch_Lightning_2019} for training. For ShapeNet models, we use $8 \times$ NVIDIA Tesla V100s for training. For other datasets, we use $8 \times$ NVIDIA Tesla A100s for training.

\subsection{Metric Definition}
To perform a quantitative comparison of our generative model on the ShapeNet dataset, we leverage the framework used in \cite{zeng2022lion} which uses the \emph{1-NNA} metric defined as follows: Given a generated set of point clouds $S_g$, a reference set of point clouds $S_r$, and a metric $D(\cdot, \cdot) : 2^{\mathbb{R}^3} \times 2^{\mathbb{R}^3} \rightarrow \mathbb{R}$ between two point clouds, the 1-NNA metric is defined as
\begin{equation}
    \text{1-NNA}(S_g, S_r) = \frac{\sum_{X \in S_g} \mathbbm{1} [N_X \in S_r] + \sum_{Y \in S_r} \mathbbm{1}[N_Y \in S_r]}{|S_g| + |S_r|},
\end{equation}
where $N_A$ is the closest point cloud to $A \in 2^{\mathbb{R}^3}$ in the set $S_g \cup S_r - \{X\}$ under the metric $D(\cdot, \cdot)$ (\textit{i.e.} the closest point cloud to $A$ in the generated and reference set not including $A$ itself), and $\mathbbm{1}[\cdot]$ is the indicator function which returns $1$ if the argument is true and $0$ otherwise.

Intuitively, the 1-NNA distance is the classification accuracy when using nearest neighbors under $D$ to determine if a point cloud was generated ($\in S_g$) or not ($\in S_r$). If the generated set is close in distribution to the reference set, then the classification accuracy should be around 50\% which is the best 1-NNA score achievable.

In our experiment, we sampled 2048 points from the surface of each shape (following~\cite{zeng2022lion}) to generate $S_g$ and $S_r$ and used the Chamfer and Earth Mover's distances as metrics $D$ to compute the 1-NNA.

\section{More results}
\label{sec:more-results}

\begin{figure*}[b]
    \centering
    \includegraphics[width=0.7\linewidth]{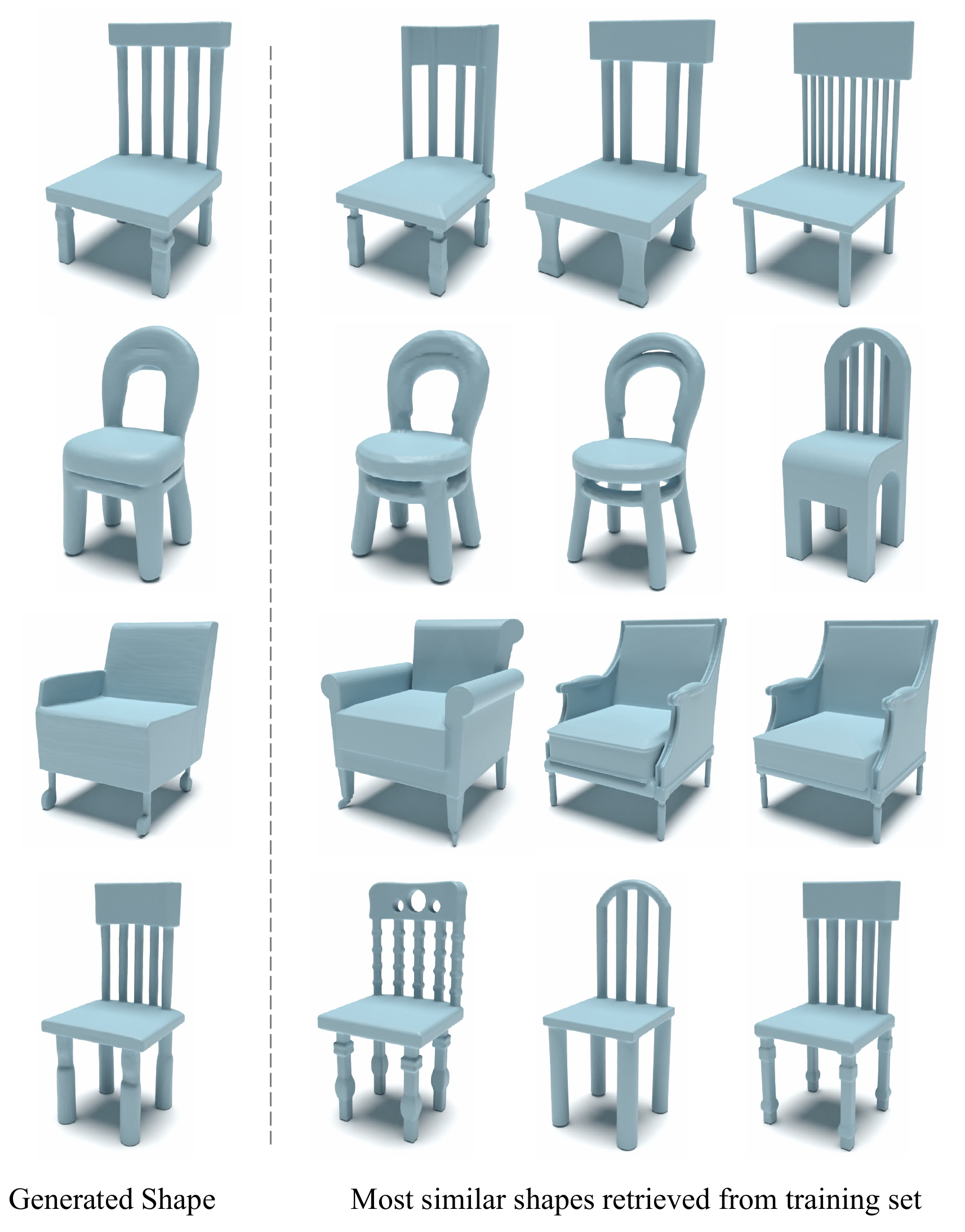}
    \caption{Shape Novelty Analysis. From our generated shape (left), we retrieve top-three most similar shapes in training set by CD distance}
    \label{fig:shapenet-supplement-1nna}
\end{figure*}

\begin{figure*}
    \centering
    \textit{"A 3D model of lion"} \\
    \includegraphics[width=0.8\linewidth]{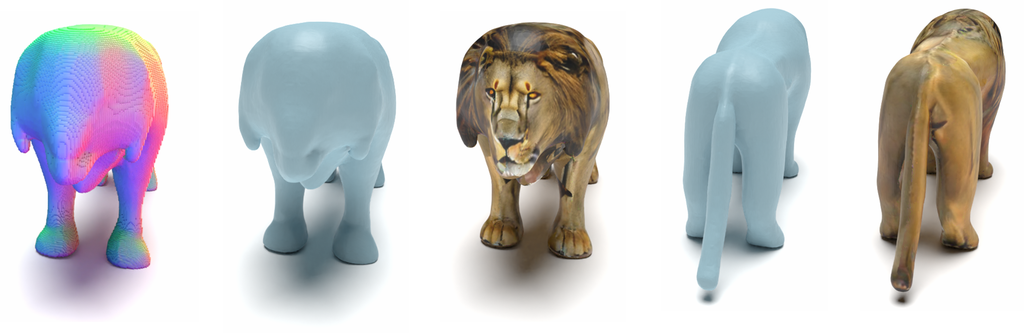} \\
    \textit{"A campfire"} \\
    \includegraphics[width=0.8\linewidth]{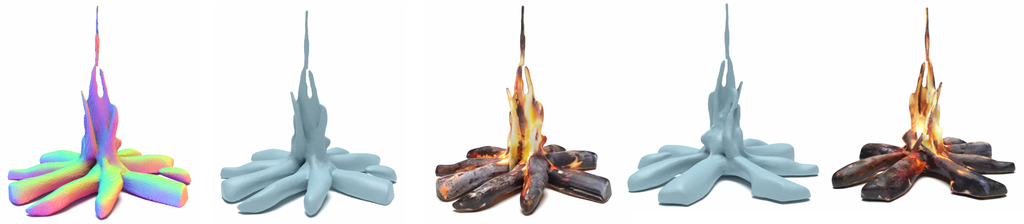} \\
    \vspace{3mm}
    \textit{"A 3D model of croissant"} \\
    \vspace{3mm}
    \includegraphics[width=0.8\linewidth]{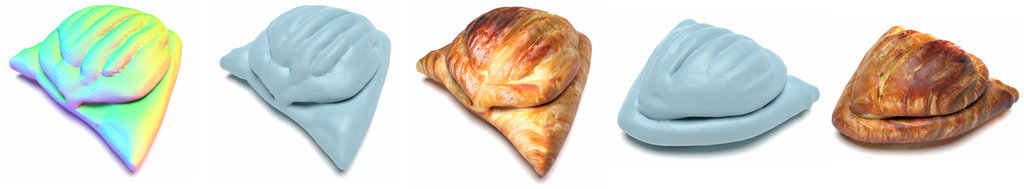} \\
    \vspace{3mm}
    \textit{"A 3D model of eagle head"} \\
    \vspace{3mm}
    \includegraphics[width=0.8\linewidth]{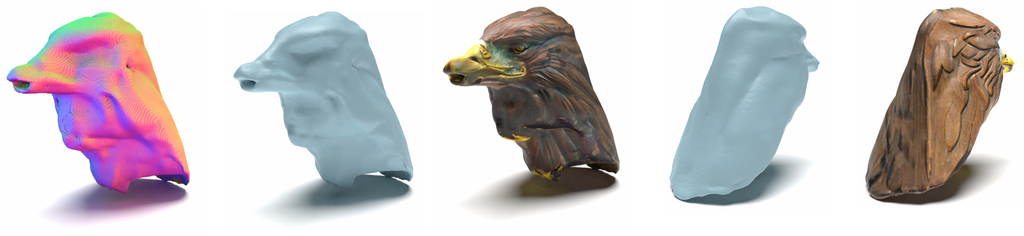} \\
    \vspace{3mm}
    \textit{"A 3D model of dragon head"} \\
    \vspace{3mm}
    \includegraphics[width=0.8\linewidth]{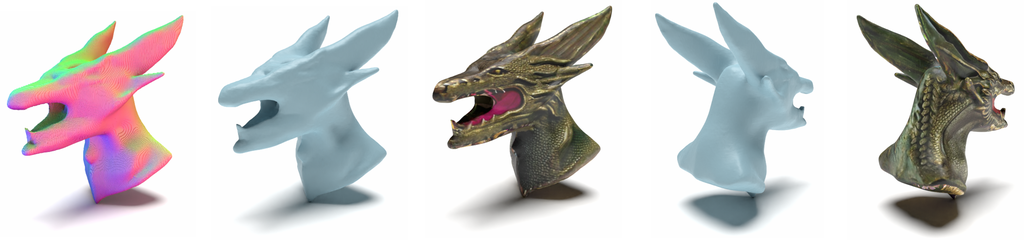} \\

    \caption{More qualitative results on text-to-3D.}
    \label{fig:objaverse-supplement-1}
\end{figure*}

\begin{figure*}
    \centering
    \textit{"A voxelized dog"} \\
    \vspace{3mm}
    \includegraphics[width=0.8\linewidth]{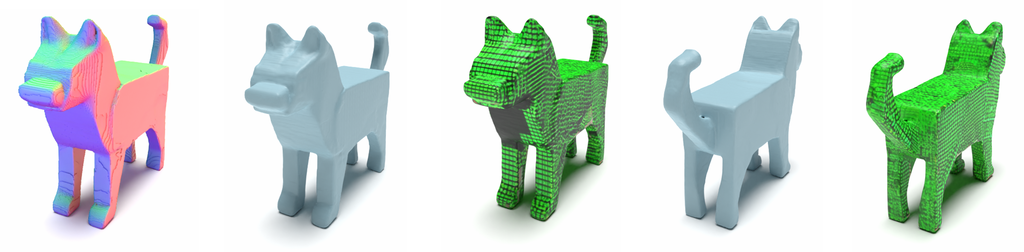} \\
    \vspace{3mm}
    \textit{"A diamond ring"} \\
    \vspace{3mm}
    \includegraphics[width=0.8\linewidth]{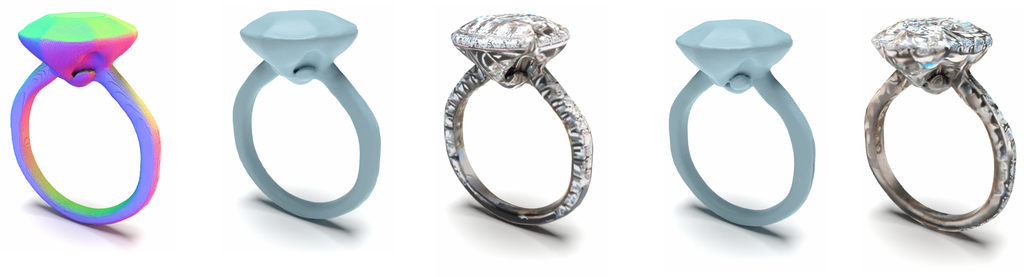} \\
    \textit{"A 3D model of cat"} \\
    \vspace{3mm}
    \includegraphics[width=0.8\linewidth]{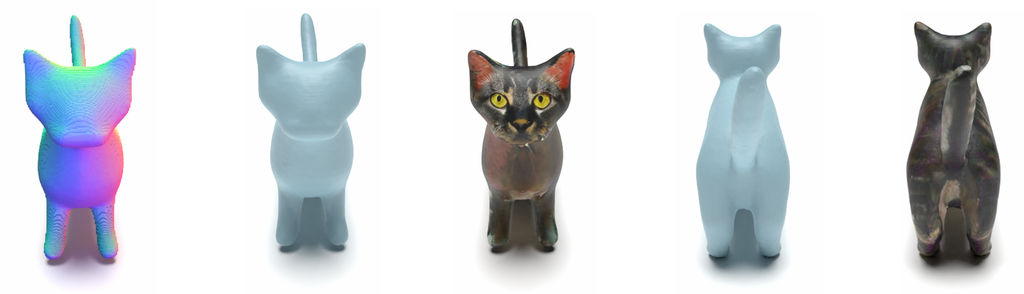} \\
    \textit{"A 3D model of duck"} \\
    \vspace{3mm}
    \includegraphics[width=0.8\linewidth]{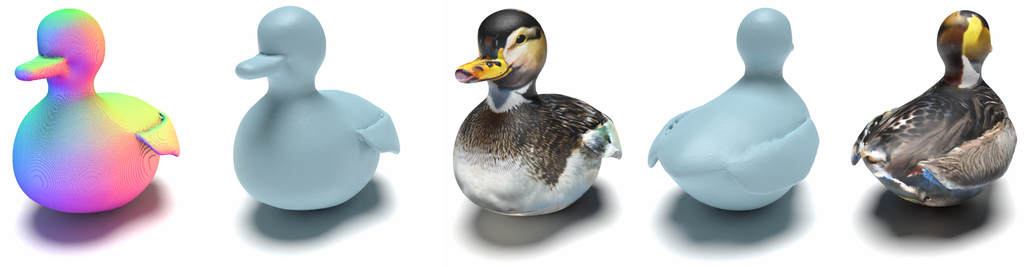} \\

    \caption{More qualitative results on text-to-3D.}
    \label{fig:objaverse-supplement-2}
\end{figure*}

\begin{figure*}
    \centering
    \textit{"A designer dress"} \\
    \vspace{3mm}
    \includegraphics[width=0.8\linewidth]{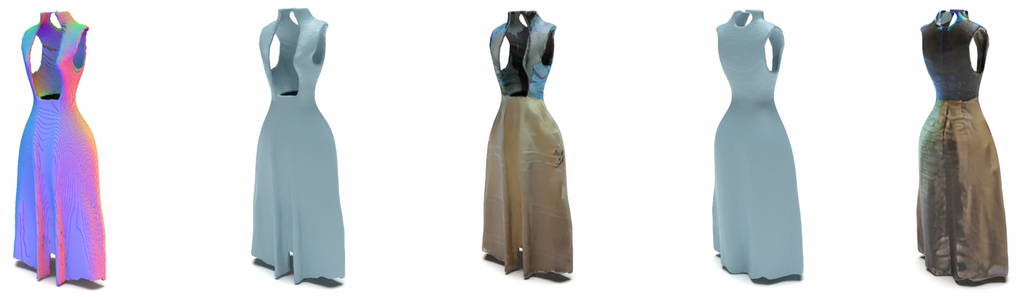} \\
    \vspace{3mm}
    \textit{"A 3D model of koala"} \\
    \vspace{3mm}
    \includegraphics[width=0.8\linewidth]{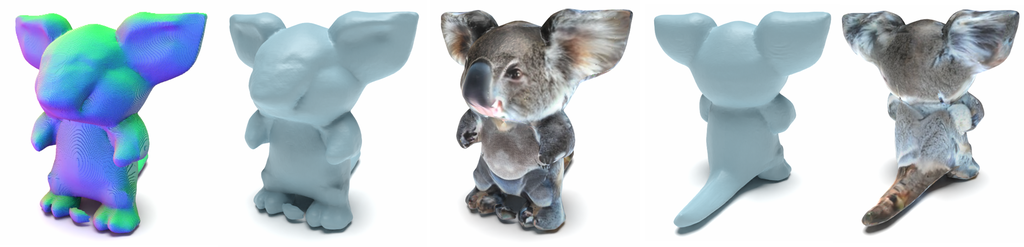} \\
    \vspace{3mm}
    \textit{"A 3D model of mushroom"} \\
    \vspace{3mm}
    \includegraphics[width=0.8\linewidth]{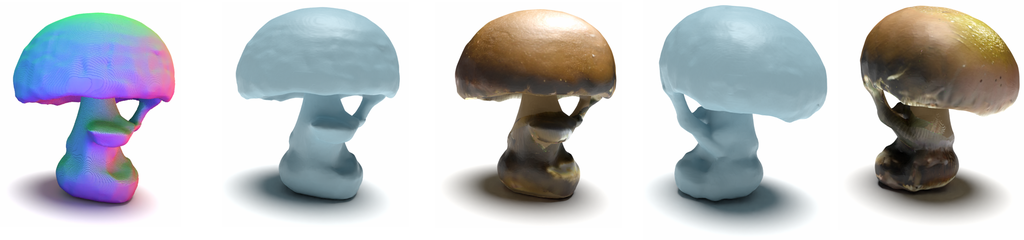} \\
    \vspace{3mm}
    \textit{"A fireplug"} \\
    \vspace{3mm}
    \includegraphics[width=0.8\linewidth]{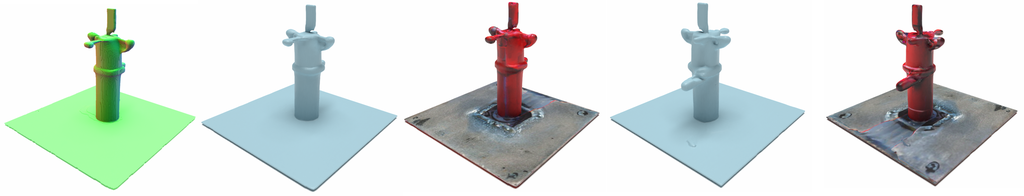} \\
    \vspace{3mm}
    \caption{More qualitative results on text-to-3D.}
    \label{fig:objaverse-supplement-3}
\end{figure*}

\begin{figure*}
    \centering
    \includegraphics[width=\linewidth]{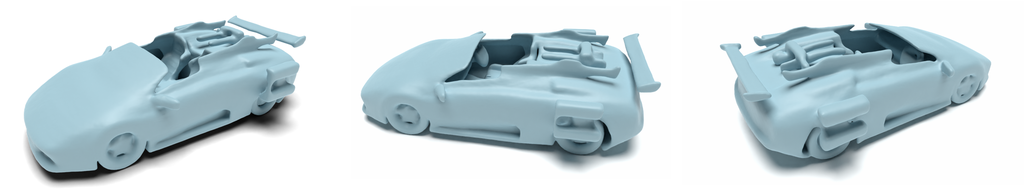}
    \includegraphics[width=\linewidth]{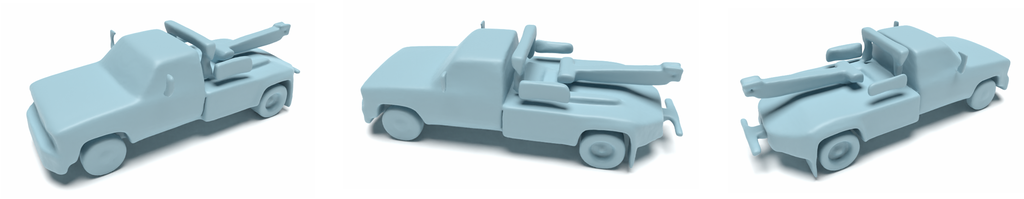}
    \includegraphics[width=\linewidth]{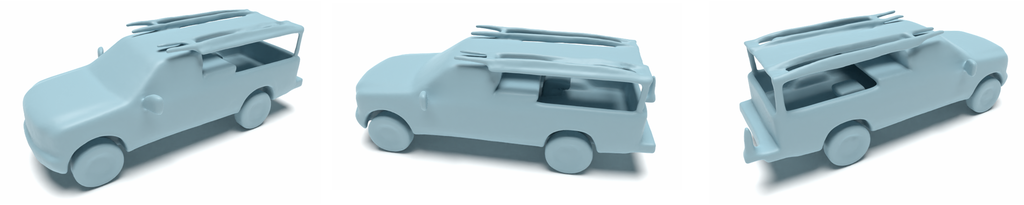}
    \includegraphics[width=\linewidth]{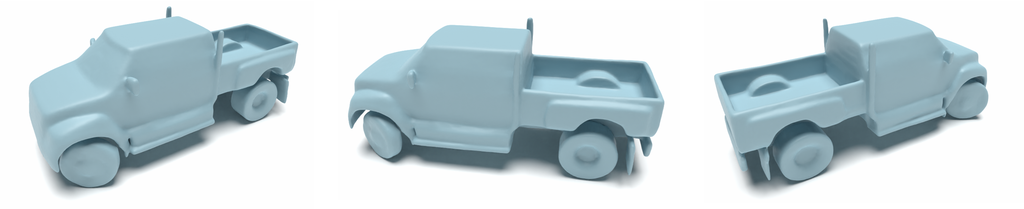}
    \includegraphics[width=\linewidth]{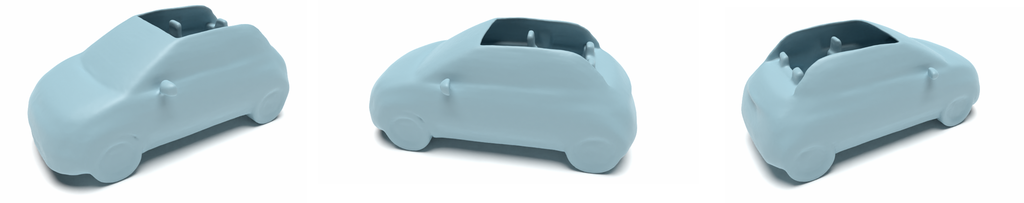}
    \includegraphics[width=\linewidth]{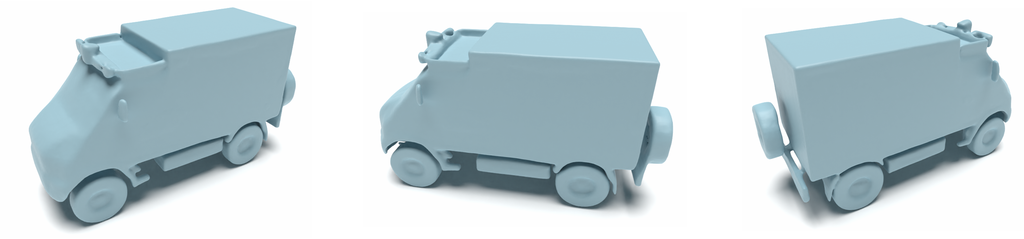}
    \caption{More qualitative results on ShapeNet Car.}
    \label{fig:shapenet-supplement-car}
\end{figure*}

\begin{figure*}
    \centering
    \includegraphics[width=0.9\linewidth]{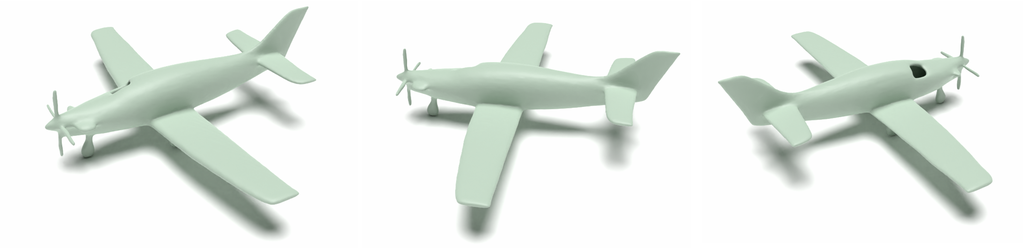}
    \includegraphics[width=0.9\linewidth]{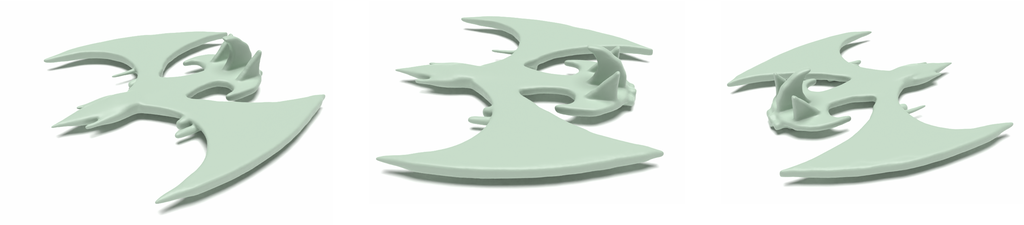}
    \includegraphics[width=0.9\linewidth]{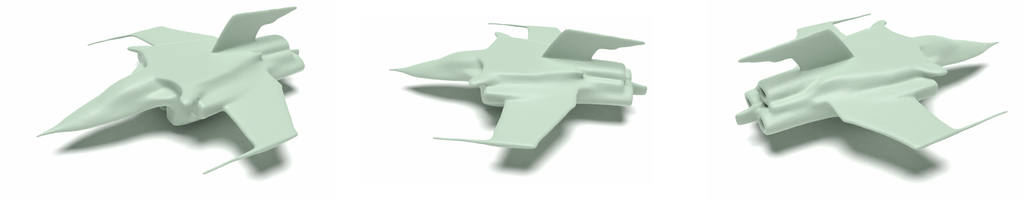}
    \includegraphics[width=0.9\linewidth]{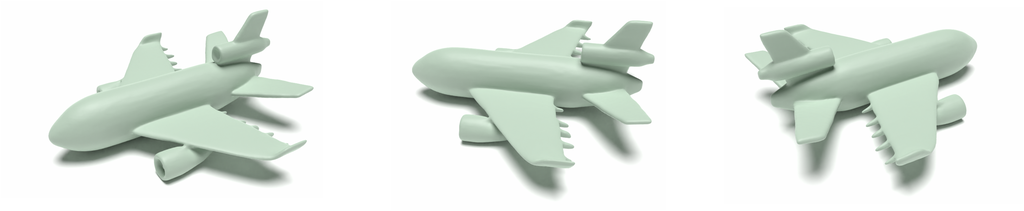}
    \includegraphics[width=0.9\linewidth]{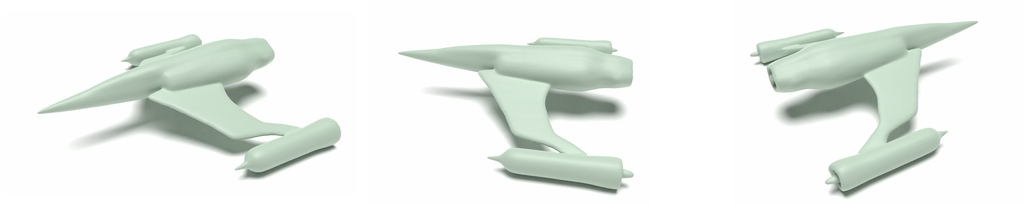}
    \includegraphics[width=0.9\linewidth]{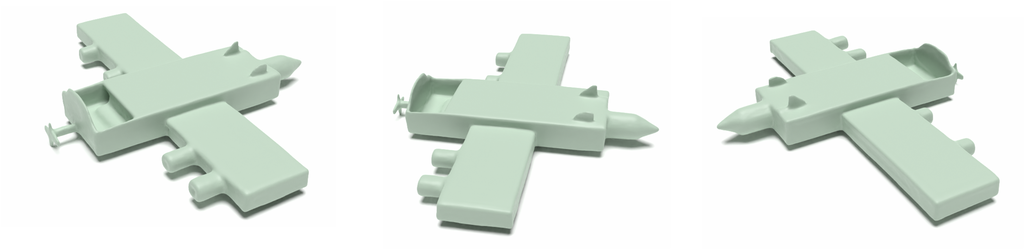}
    \caption{More qualitative results on ShapeNet Airplane.}
    \label{fig:shapenet-supplement-plane}
\end{figure*}

\begin{figure*}
\centering
\begin{tabular}{c@{\hspace{3mm}}c}
\includegraphics[width=0.48\linewidth]{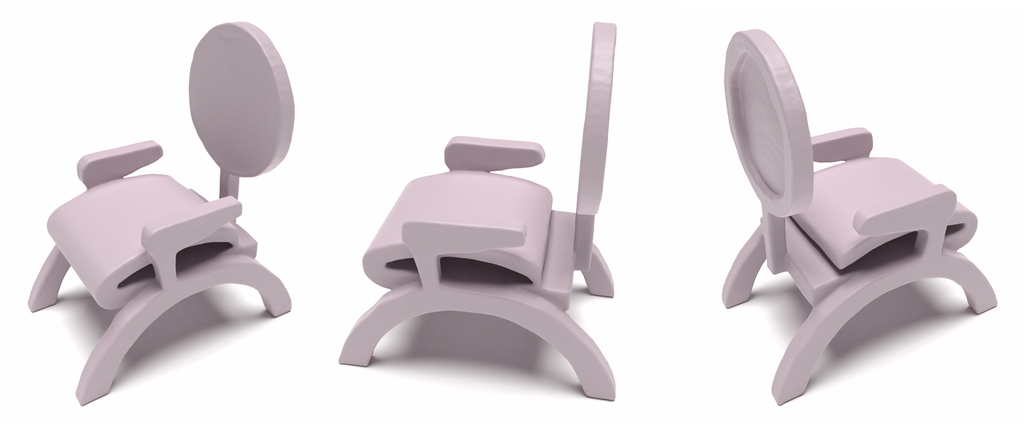}&
\includegraphics[width=0.48\linewidth]{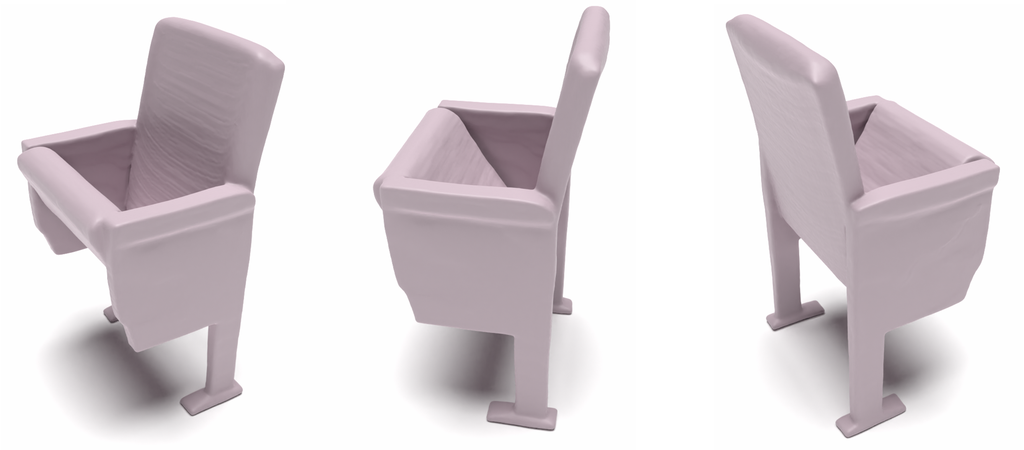}\\
\includegraphics[width=0.48\linewidth]{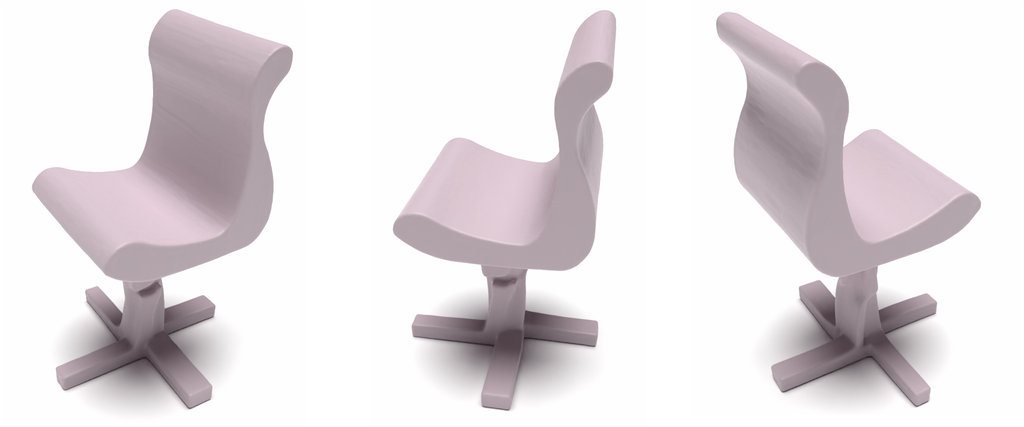}&
\includegraphics[width=0.48\linewidth]{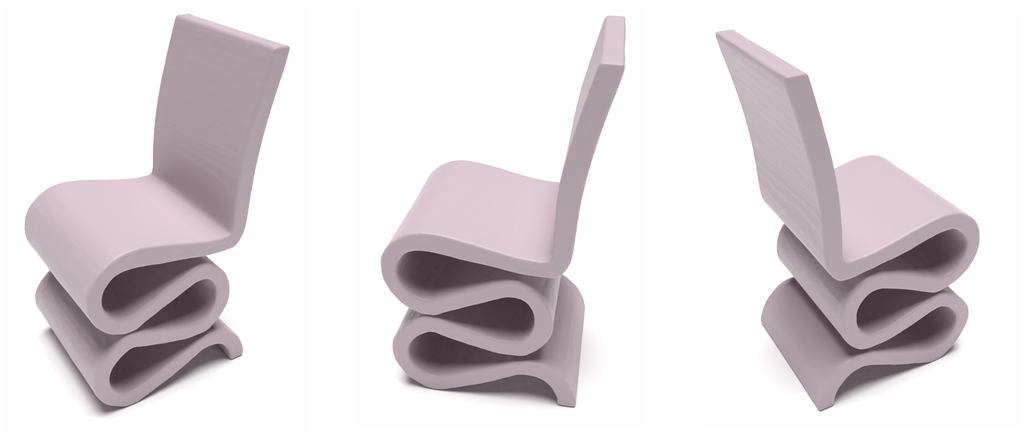}\\
\includegraphics[width=0.48\linewidth]{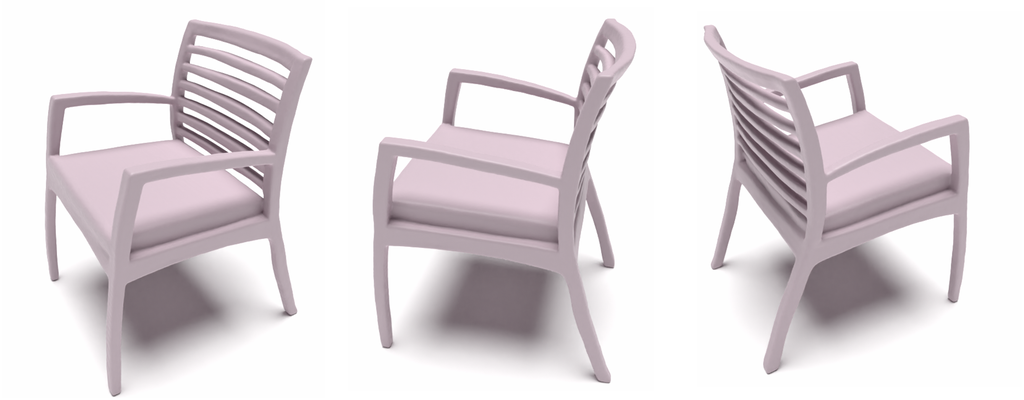}&
\includegraphics[width=0.48\linewidth]{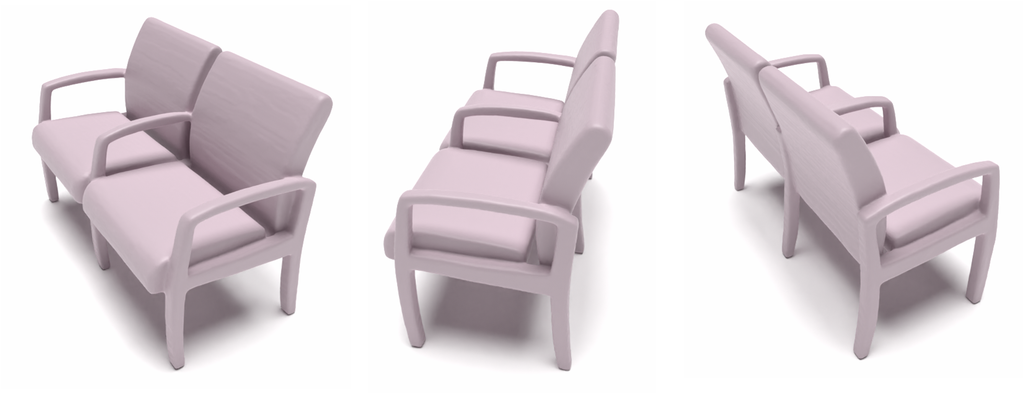}\\
\includegraphics[width=0.48\linewidth]{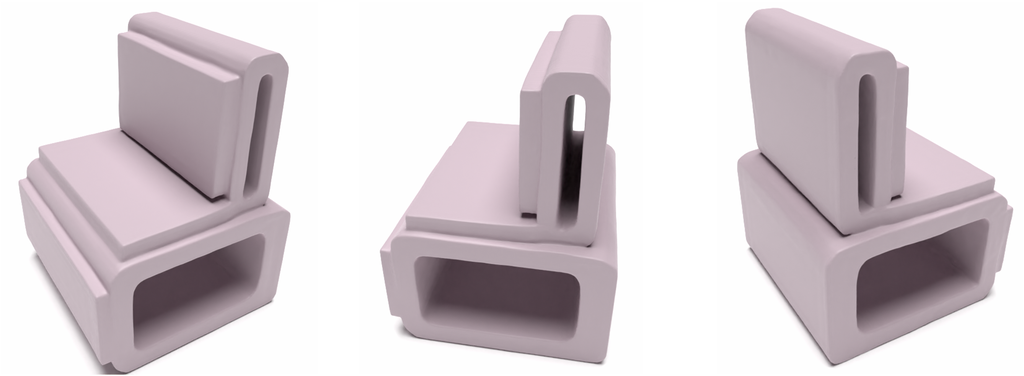}&
\includegraphics[width=0.48\linewidth]{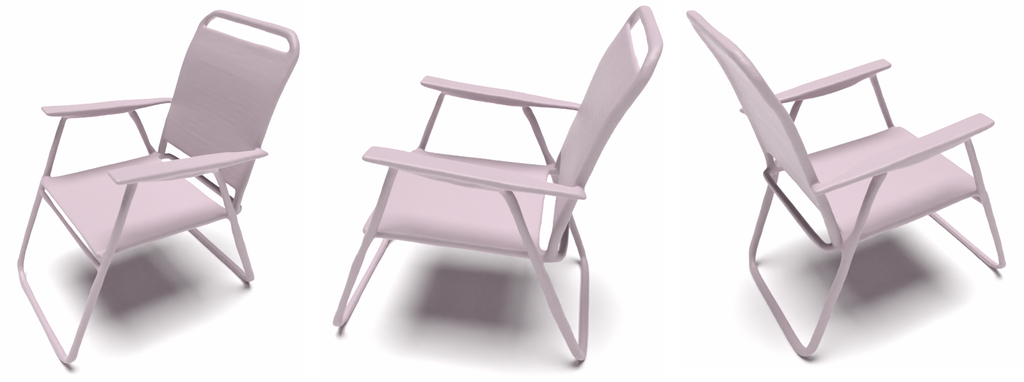}\\
\includegraphics[width=0.48\linewidth]{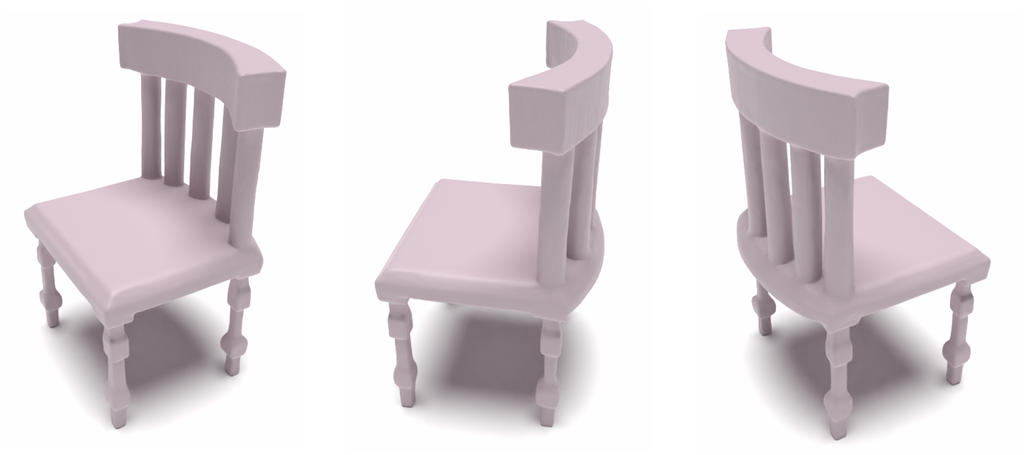}&
\includegraphics[width=0.48\linewidth]{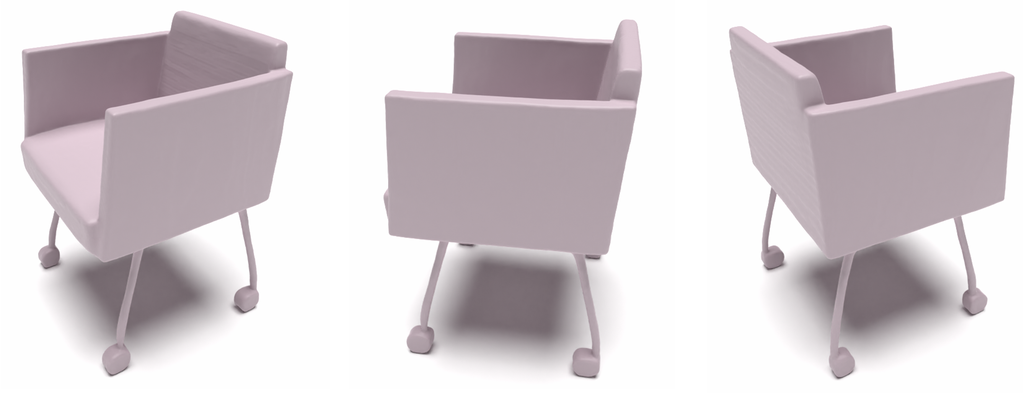}\\
\includegraphics[width=0.48\linewidth]{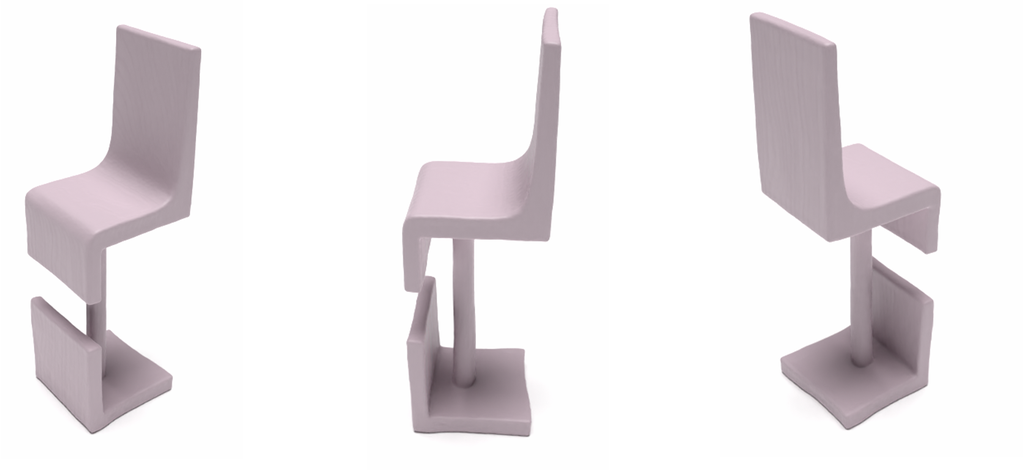}&
\includegraphics[width=0.48\linewidth]{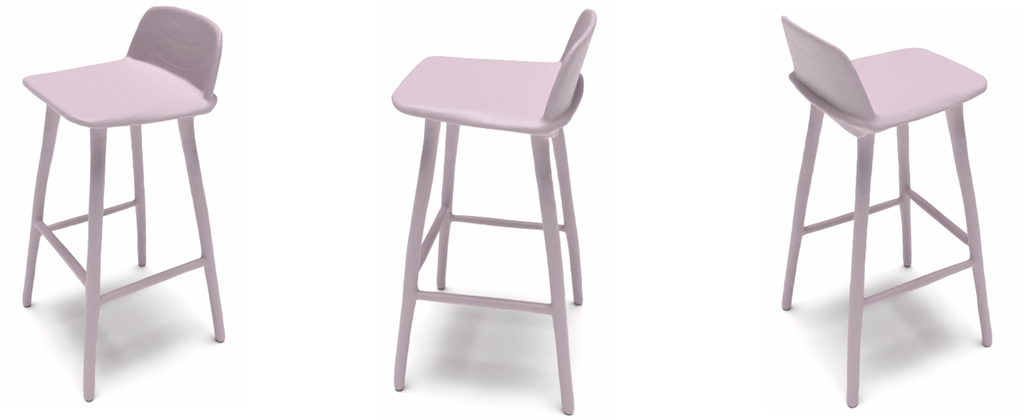}\\
\end{tabular}
\caption{More qualitative results on ShapeNet Chair.}
\label{fig:shapenet-supplement-chair}
\end{figure*}

\begin{figure*}
\centering
\begin{tabular}{cc}
\includegraphics[width=0.48\linewidth]{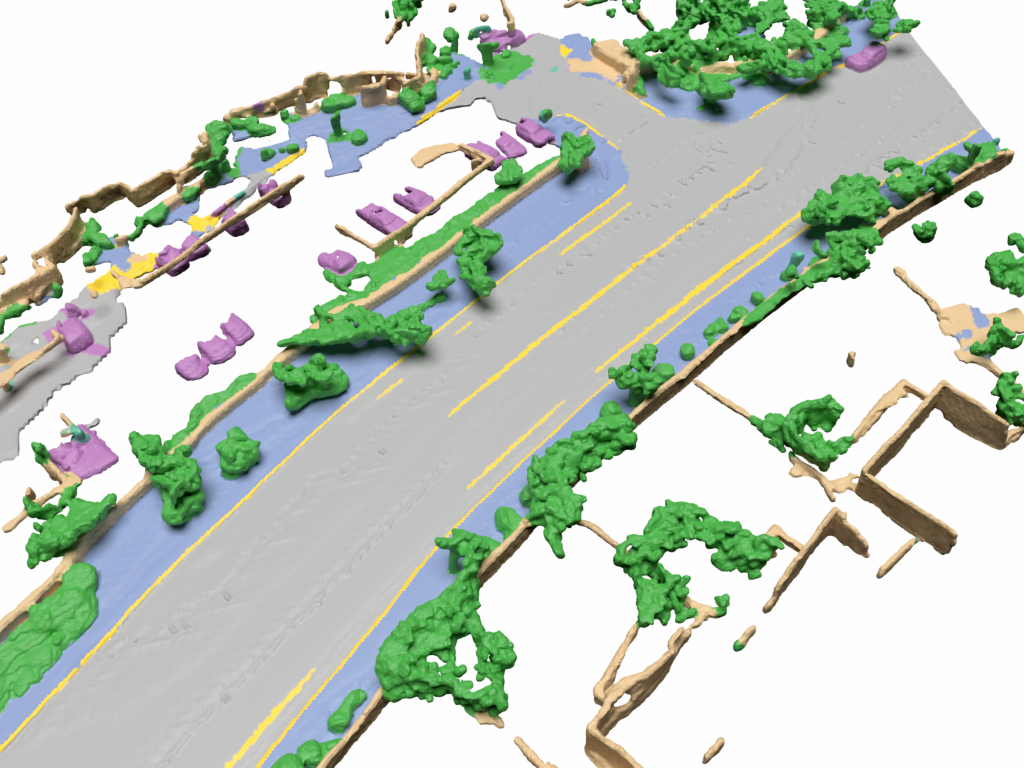}&
\includegraphics[width=0.48\linewidth]{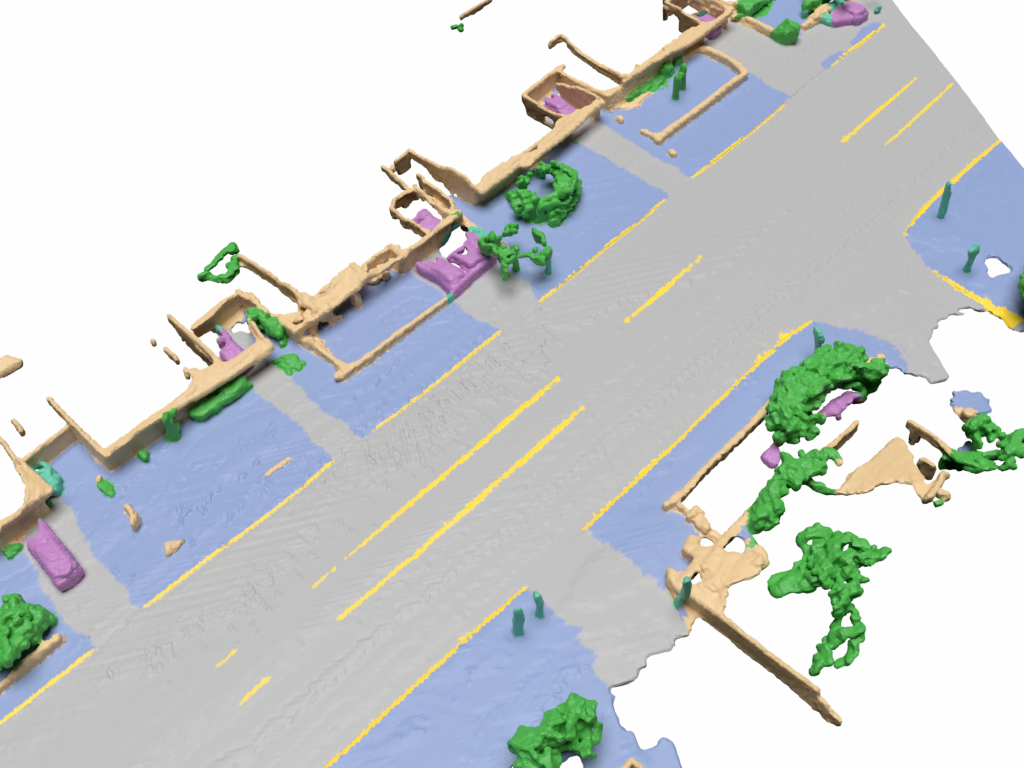}\\
\includegraphics[width=0.48\linewidth]{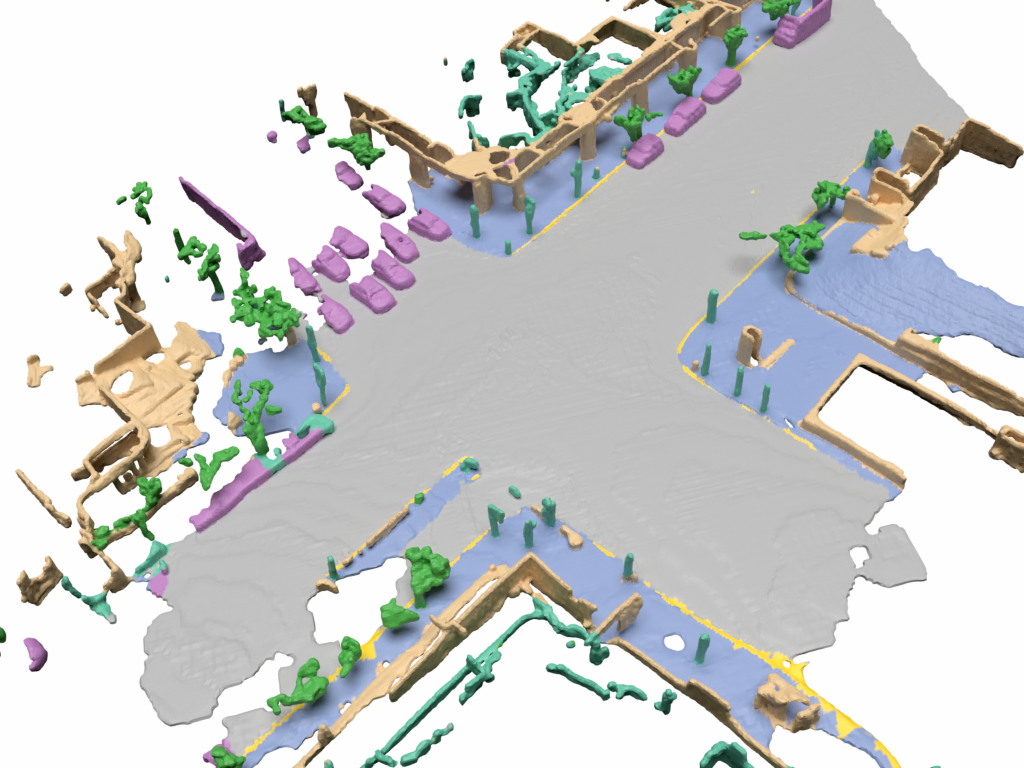}&
\includegraphics[width=0.48\linewidth]{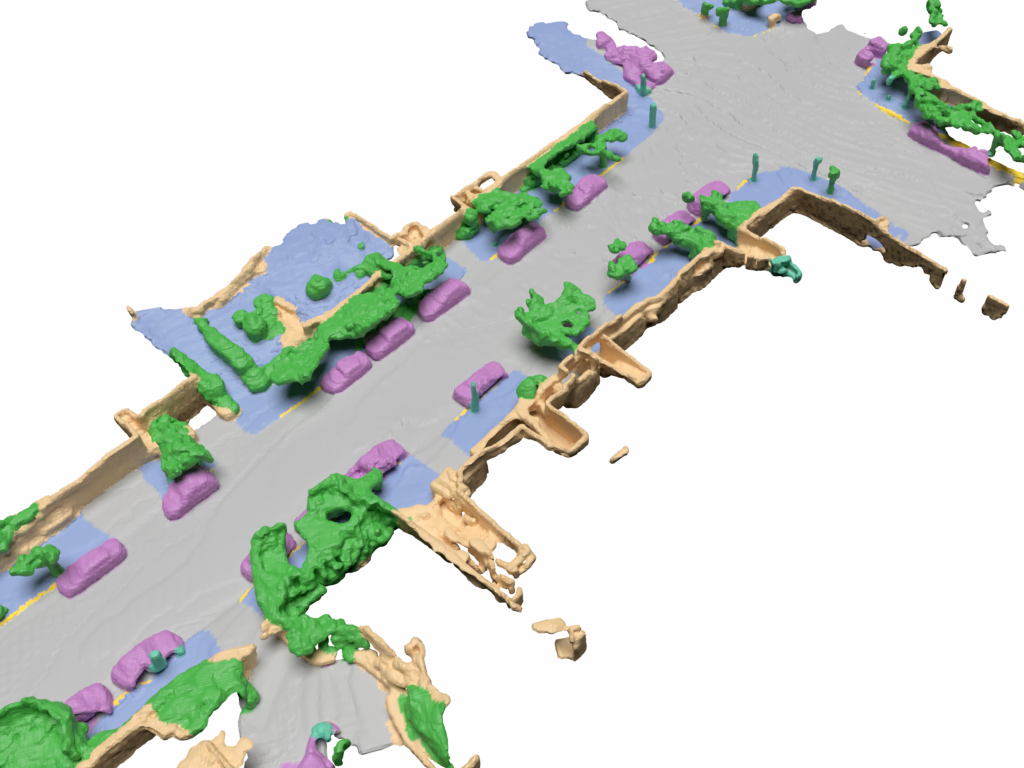}\\
\includegraphics[width=0.48\linewidth]{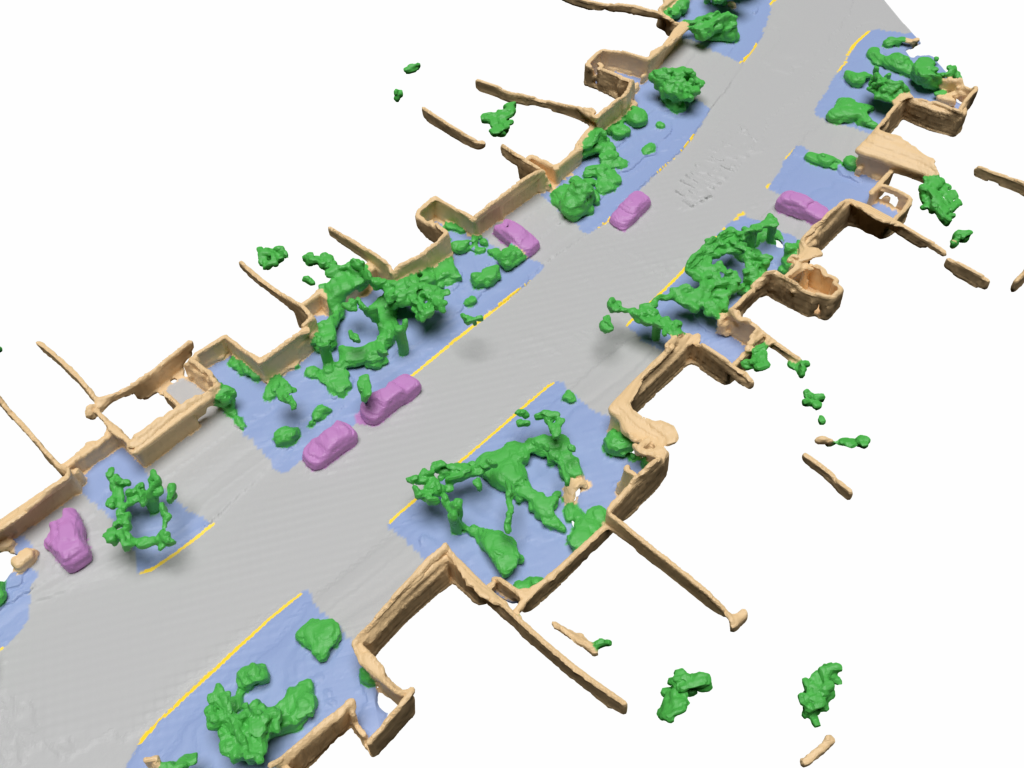}&
\includegraphics[width=0.48\linewidth]{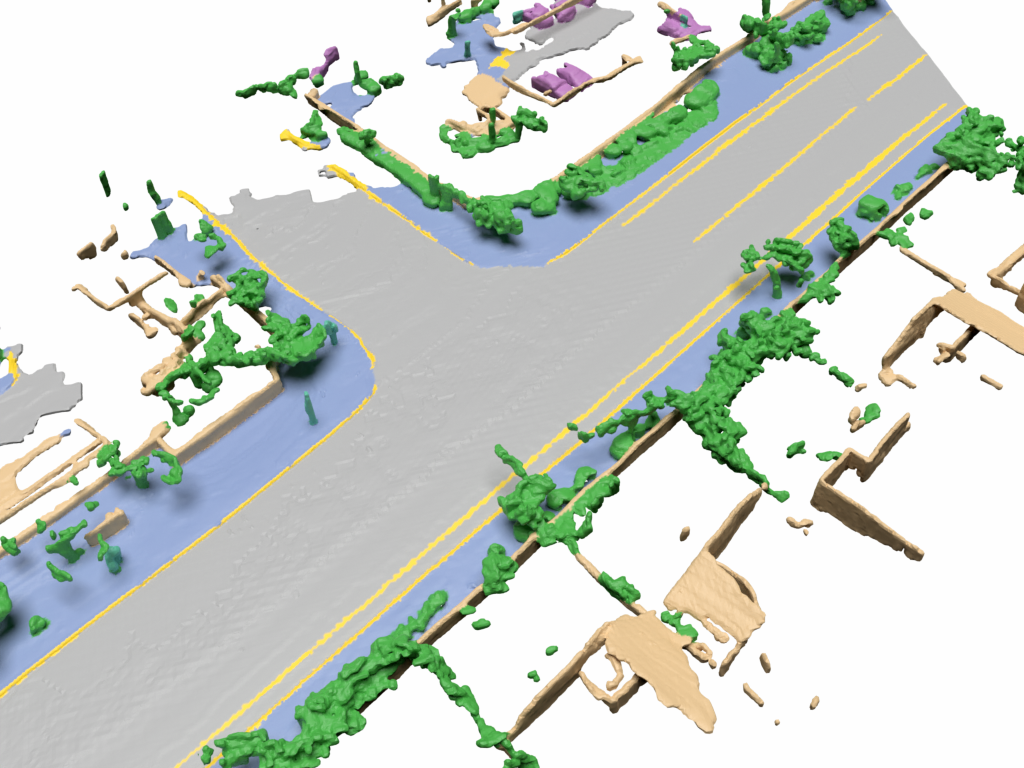}\\
\end{tabular}
\caption{More qualitative results on Waymo.}
\label{fig:waymo-supplement-1}
\end{figure*}

\begin{figure*}
\centering
\begin{tabular}{cc}
\includegraphics[width=0.48\linewidth]{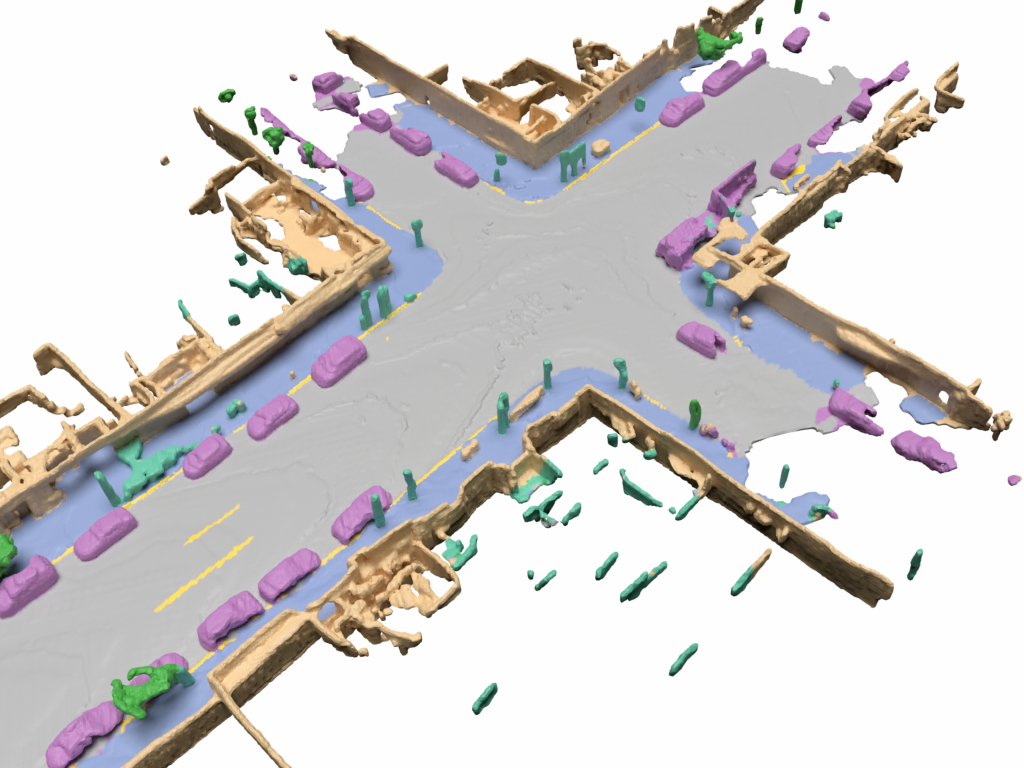}&
\includegraphics[width=0.48\linewidth]{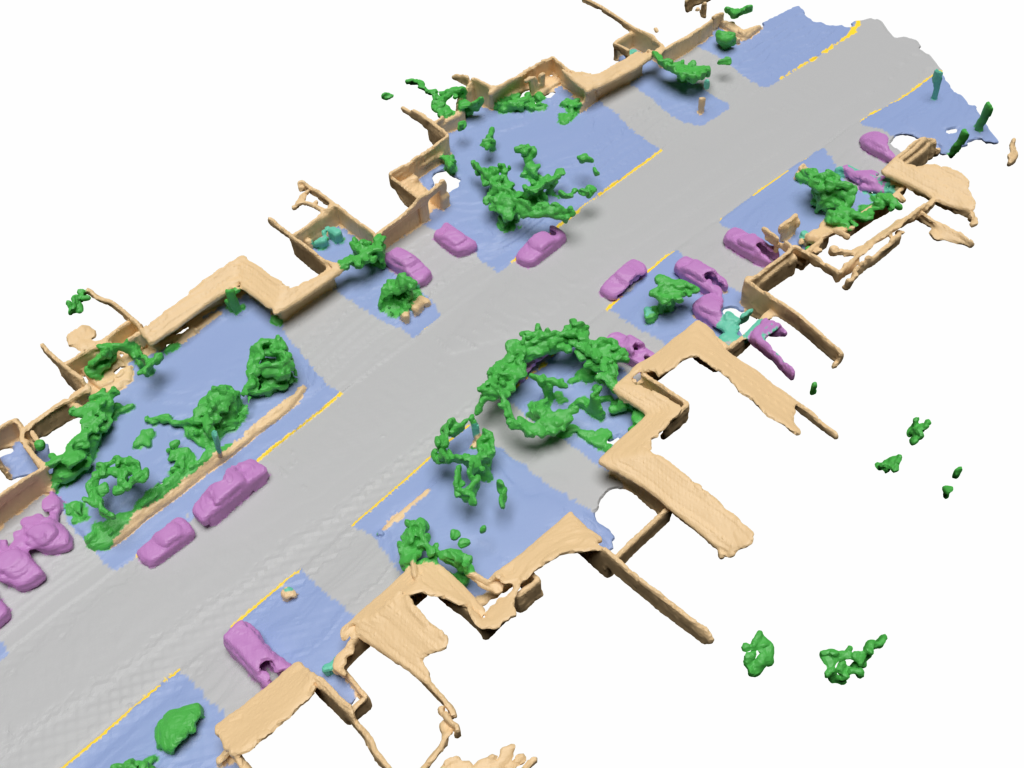}\\
\includegraphics[width=0.48\linewidth]{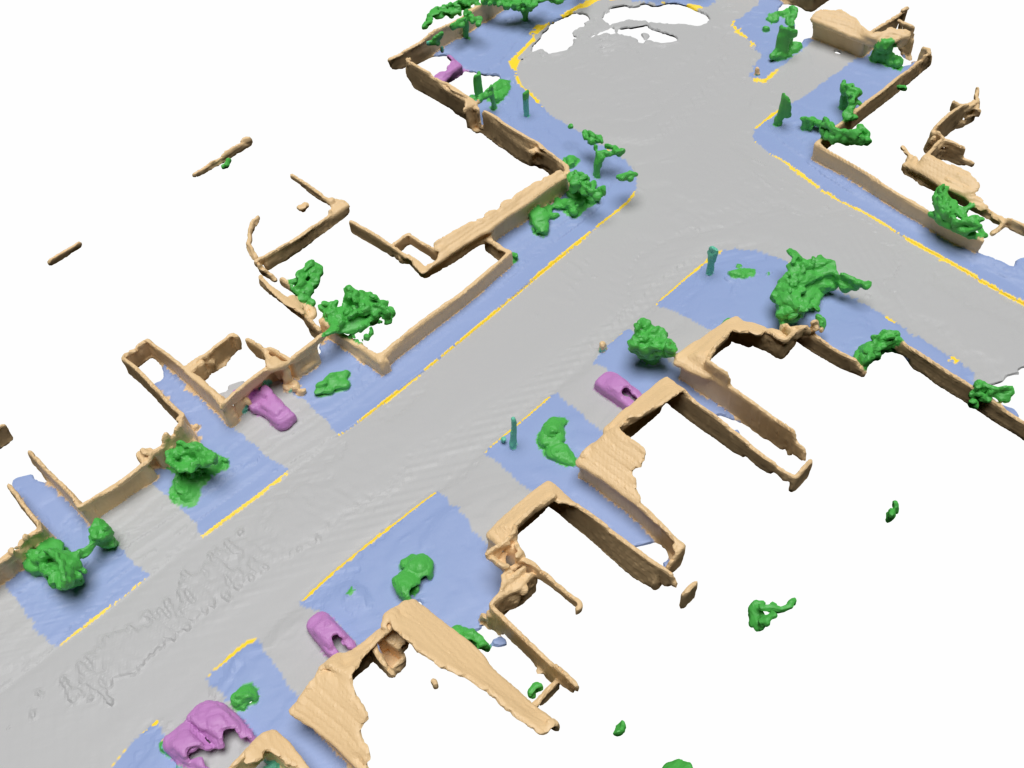}&
\includegraphics[width=0.48\linewidth]{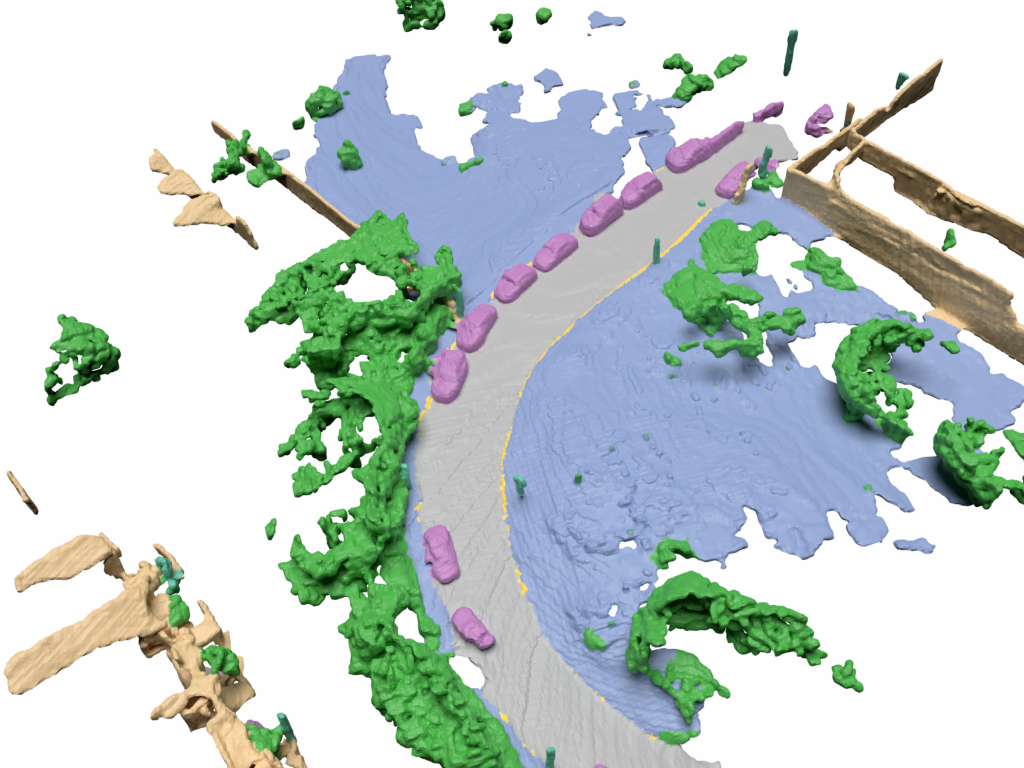}\\
\includegraphics[width=0.48\linewidth]{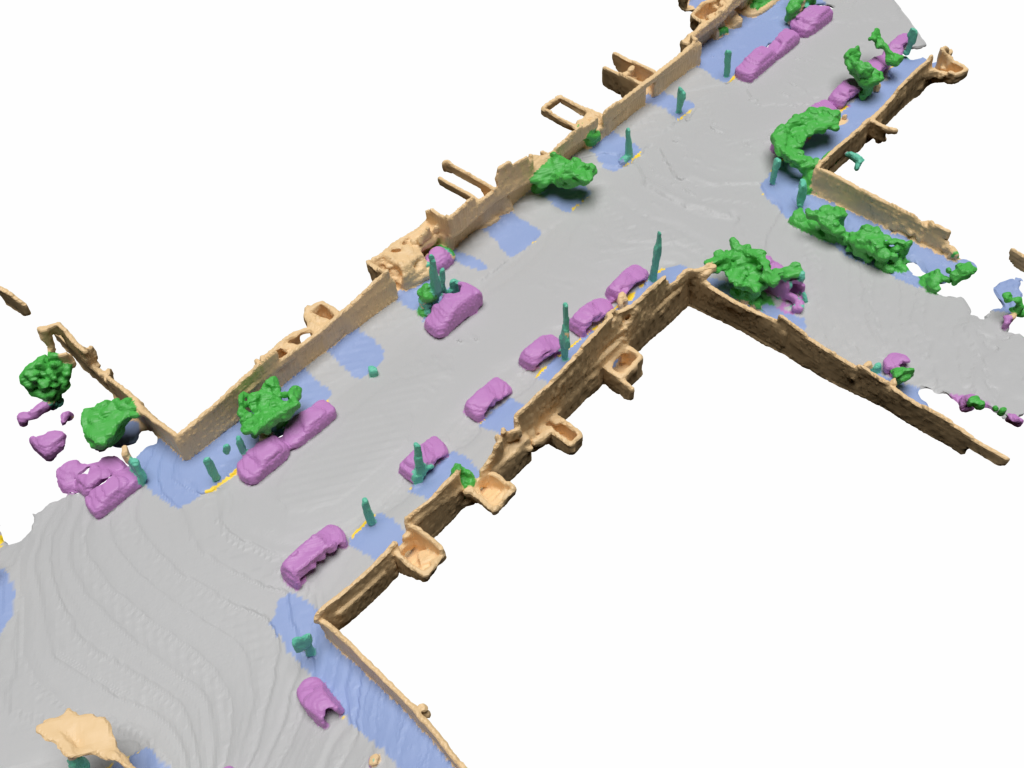}&
\includegraphics[width=0.48\linewidth]{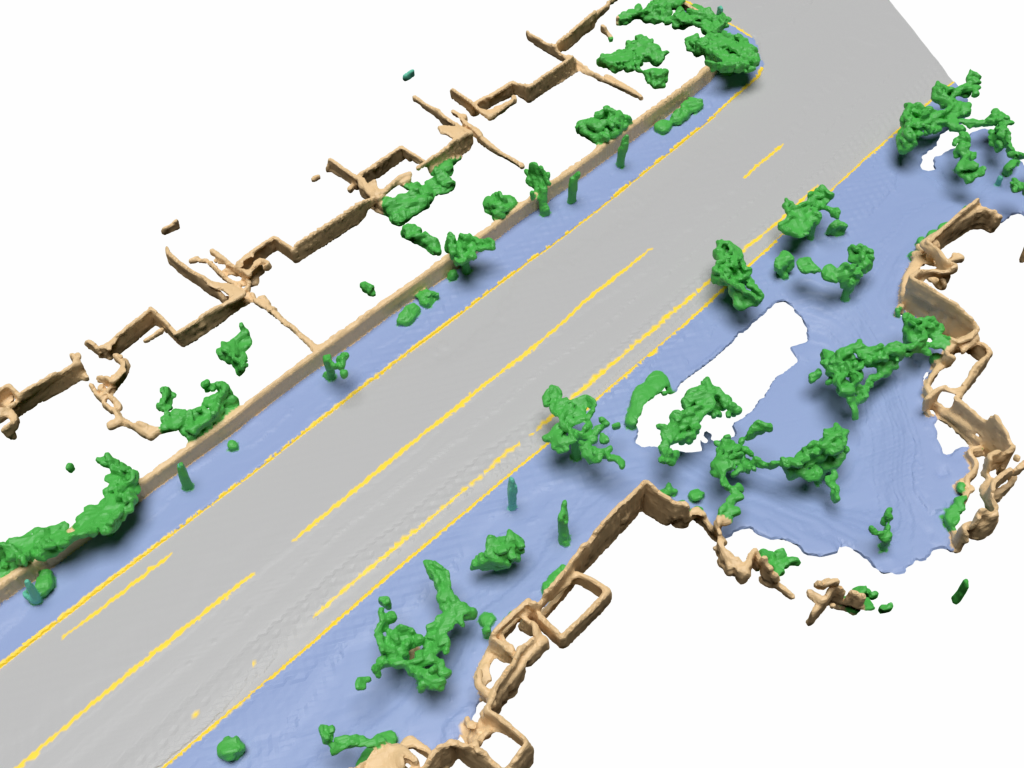}\\
\end{tabular}
\caption{More qualitative results on Waymo.}
\label{fig:waymo-supplement-2}
\end{figure*}

\begin{figure*}
\centering
\begin{tabular}{cc}
\includegraphics[width=0.48\linewidth]{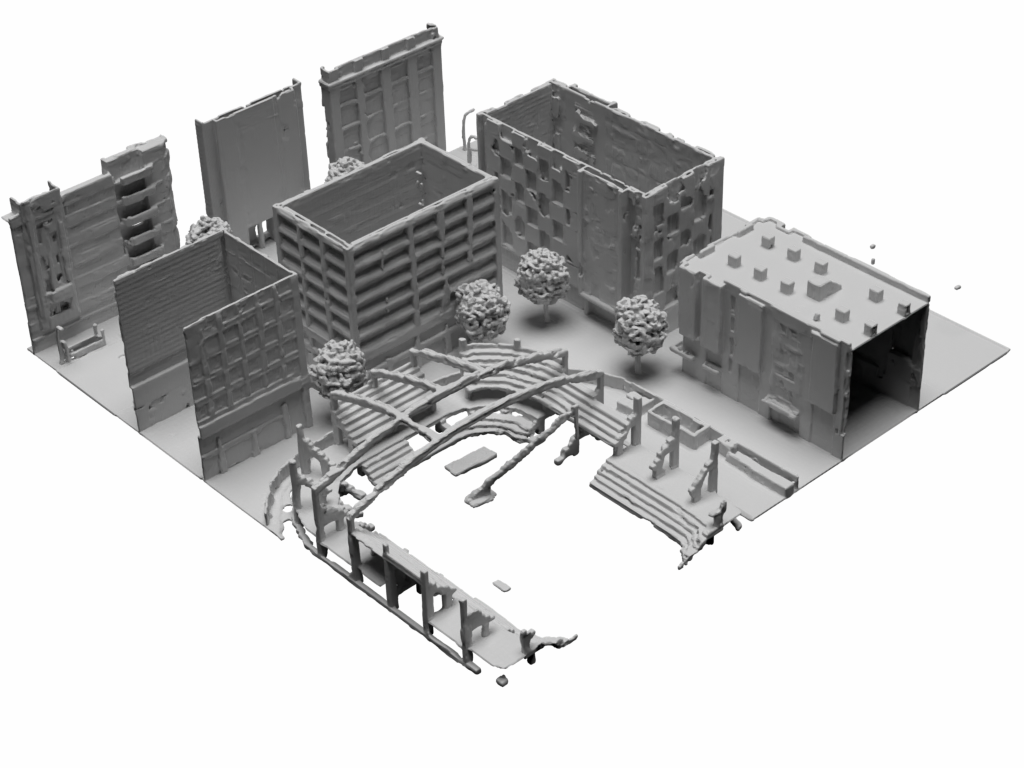}&
\includegraphics[width=0.48\linewidth]{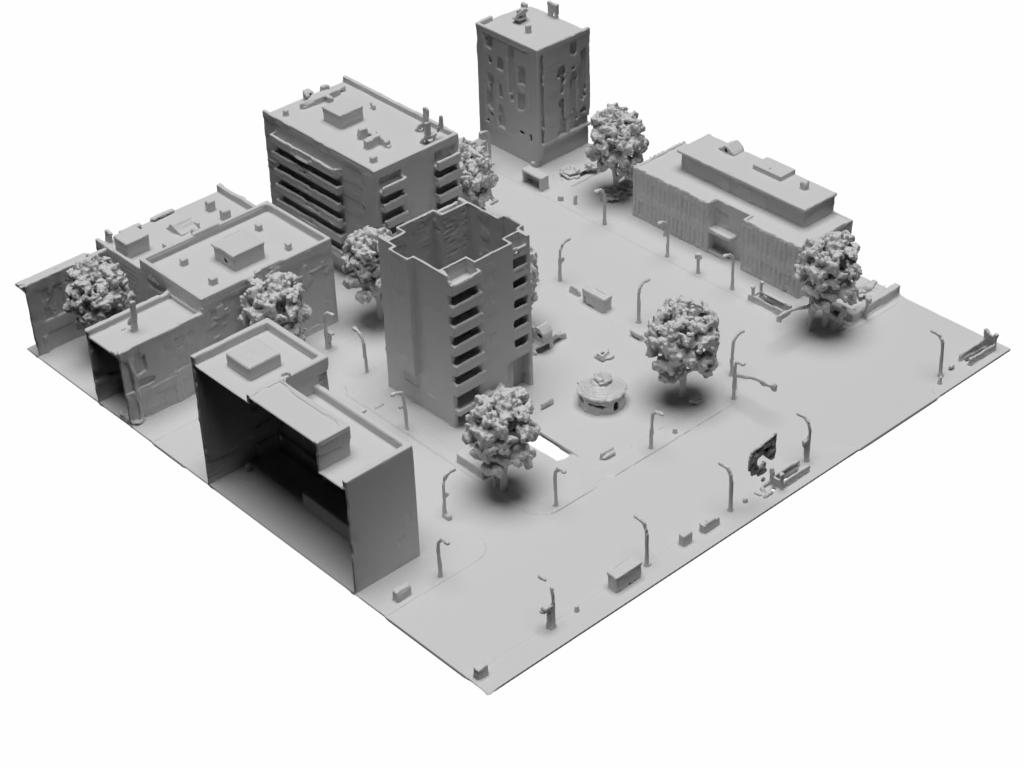}\\
\includegraphics[width=0.48\linewidth]{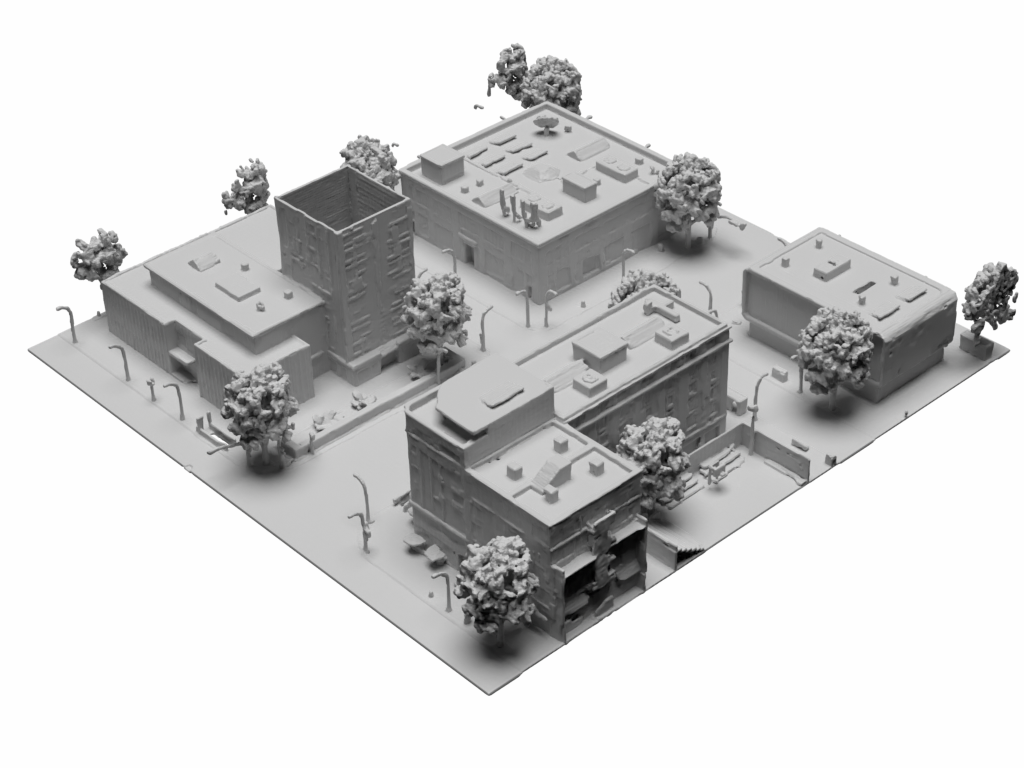}&
\includegraphics[width=0.48\linewidth]{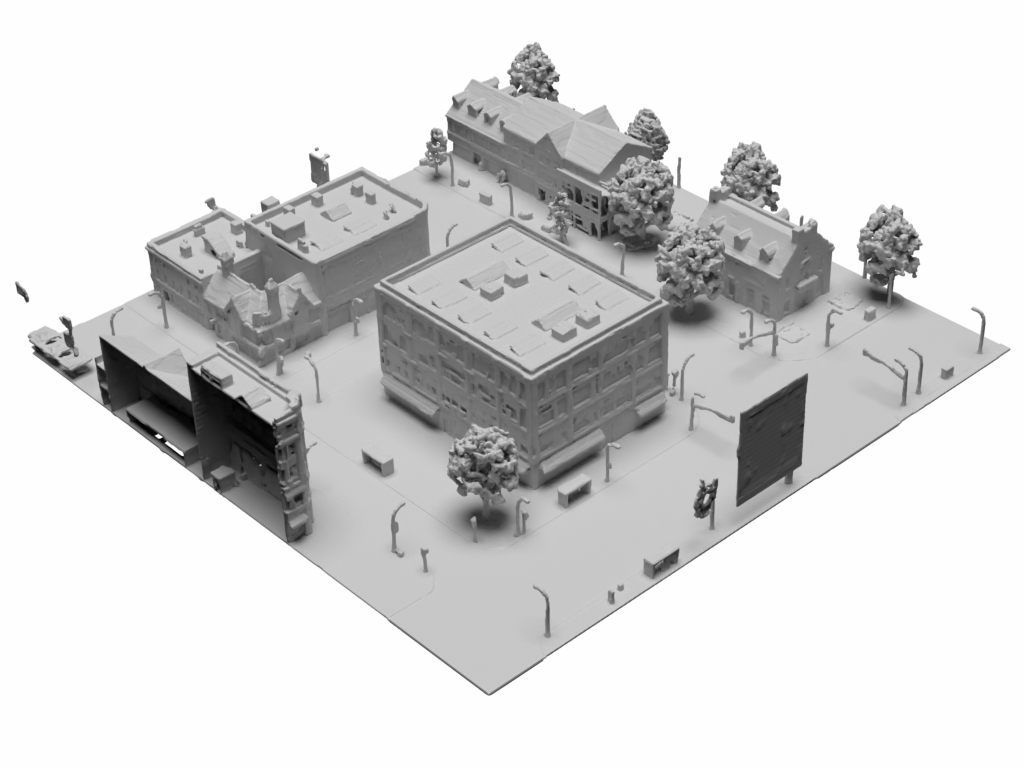}\\
\includegraphics[width=0.48\linewidth]{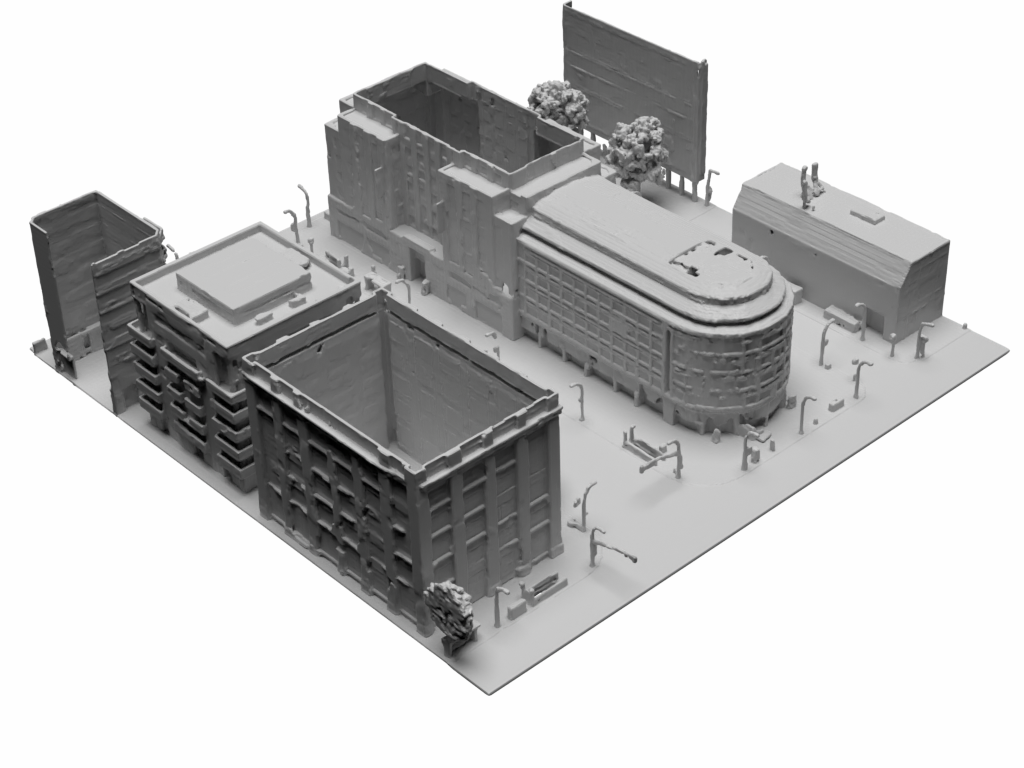}&
\includegraphics[width=0.48\linewidth]{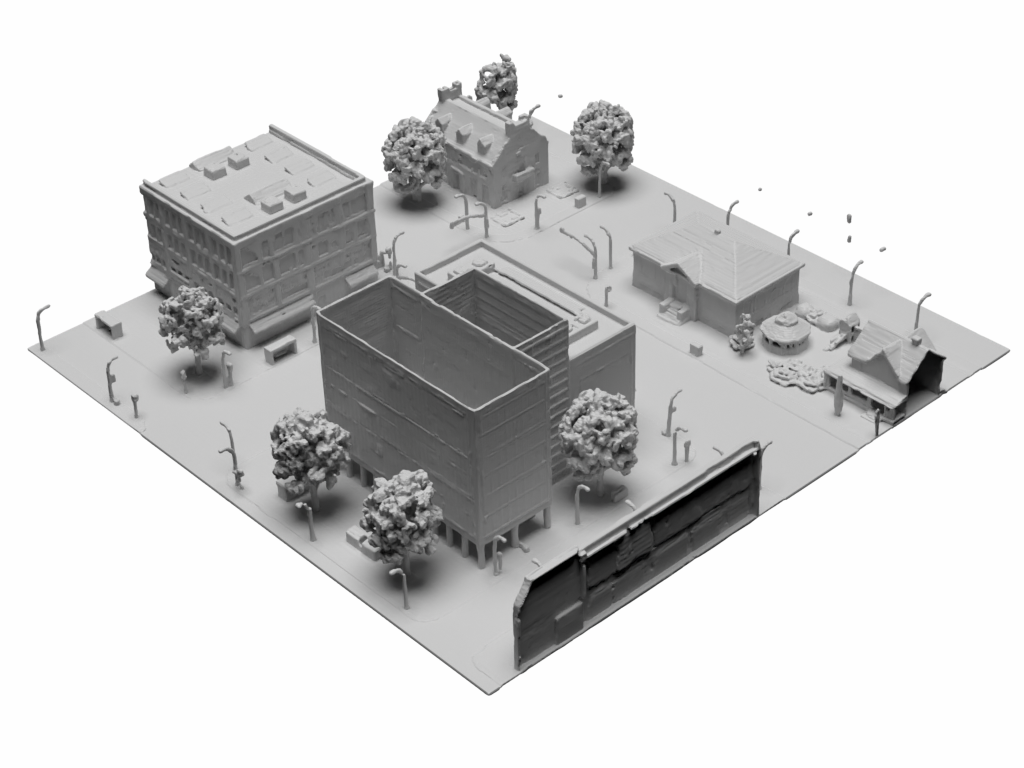}\\
\end{tabular}
\caption{More qualitative results on Karton City.}
\label{fig:turbo-squid-supplement}
\end{figure*}

In this section, we provide more qualitative results on all datasets.
First, we show more text-to-3D results on Objaverse in \cref{fig:objaverse-supplement-1,fig:objaverse-supplement-2,fig:objaverse-supplement-3}. Then, we show more results on ShapeNet in \cref{fig:shapenet-supplement-1nna,fig:shapenet-supplement-car,fig:shapenet-supplement-plane,fig:shapenet-supplement-chair}.
Despite the high quality of our generated shapes, we show that our model does not overfit the training samples and is able to generate novel shapes in \cref{fig:shapenet-supplement-1nna} by retrieving the most similar shapes in the training set given the generated samples.
Furthermore, we show more results on Waymo in \cref{fig:waymo-supplement-1,fig:waymo-supplement-2}.
Finally, we show more results on Karton City in \cref{fig:turbo-squid-supplement}.

\clearpage

\clearpage

\clearpage

\clearpage

\clearpage

\clearpage

\clearpage

\clearpage

\clearpage

\clearpage

\begin{table}[t]
\begin{center}
\scalebox{0.9}{
\begin{tabular}{lcccccccc}
\toprule
 & ShapeNet & ShapeNet  & Objaverse & Objaverse & Waymo  & Waymo \\
 & $16^3 \rightarrow 128^3$ & $128^3 \rightarrow 512^3$ & $16^3 \rightarrow 128^3$ & $128^3 \rightarrow 512^3$ &  $32^3 \rightarrow 256^3$ & $256^3 \rightarrow 1024^3$ \\
\midrule
Model Size & 59.6M & 3.8M & 236M & 14.9M & 59.4M & 3.8M \\
Base Channels & 64 & 32 & 128 & 64 & 64 & 32 \\
Channels Multiple & 1,2,4,8 & 1,2,4 & 1,2,4,8 & 1,2,4 & 1,2,4,8 & 1,2,4 \\
Latent Dim & 16 & 8 & 16 & 8 & 16 & 8 \\
Batch Size & 16 & 32 & 32 & 32 & 32 & 32 \\
Epochs & 100 & 100 & 25 & 10 & 50 & 50\\
Learning Rate & \multicolumn{6}{c}{1e-4} \\
\bottomrule
\end{tabular}
}
\end{center}
\caption{ \textbf{Hyperparameters for VAE.}  For the Karton City dataset, we used the same hyperparameters as the Waymo dataset.}
\label{table:train_vae}
\end{table}

\begin{table}[t]
\begin{center}
\scalebox{0.9}{
\begin{tabular}{lcccccccc}
\toprule
 & ShapeNet - $16^3$ & ShapeNet - $128^3$ & Objaverse - $16^3$ & Objaverse - $128^3$ & Waymo - $32^3$  & Waymo - $256^3$  \\
\midrule
Diffusion Steps & \multicolumn{6}{c}{1000} \\
Noise Schedule & \multicolumn{6}{c}{linear} \\
Model Size & 691M & 79.6M & 1.5B & 79.6M & 702M & 76.6M \\
Base Channels & 192 & 64 & 256 & 64 & 192 & 64  \\
Depth & \multicolumn{6}{c}{2} \\
Channels Multiple & 1,2,4,4 & 1,2,2,4 & 1,2,4,4 & 1,2,2,4 & 1,2,4,4 & 1,2,2,4 \\
Heads & \multicolumn{6}{c}{8} \\
Attention Resolution & 16,8,4 & 32,16 & 16,8,4 & 32,16 & 16,8 & 32  \\
Dropout & 0.1 & 0.0 & 0.0 & 0.0 & 0.0 & 0.0 \\
Batch Size & 256 & 256 & 512 & 128 & 512 & 256 \\
Iterations & varies* & 15K & 95K & 80K & 40K & 20K \\
Learning Rate & \multicolumn{6}{c}{5e-5} \\
\bottomrule
\end{tabular}
}
\end{center}
\caption{ \textbf{Hyperparameters for voxel latent diffusion models.} *We train our model with 25K iterations for ShapeNet Airplane, 45K iterations for ShapeNet Car, and 35K iterations for ShapeNet Chair. For the Karton City dataset, we used the same hyperparameters as the Waymo dataset and trained the models to converge.}
\label{table:train}
\end{table}

\clearpage

\end{document}